\documentclass[acmsmall]{acmart}

\usepackage{latexsym}
\usepackage[utf8]{inputenc}
\usepackage{bbm}
\usepackage{microtype}

\usepackage{graphicx}

\usepackage{tabularx} 

\usepackage{booktabs}
\usepackage{multirow}
\usepackage{graphicx}

\usepackage{algorithm}
\usepackage{algorithmic}

\usepackage{newfloat}
\usepackage{listings}

\usepackage{subfigure}
\usepackage{mathrsfs}
\usepackage{lineno}
\usepackage{xcolor}
\usepackage{color, soul}
\usepackage[inline]{enumitem}
\usepackage{acronym}

\usepackage{tikz}
\usetikzlibrary{calc,shapes.callouts}
\usepackage{tcolorbox}

\usepackage[skip=2pt]{caption}

\usepackage{amsmath}
\usepackage{arydshln}
\usepackage{fontawesome}
\AtBeginDocument{%
  }

\acrodef{IR}{information retrieval}
\acrodef{LLM}{large language model}
\acrodef{RLAIF}{reinforcement learning from AI feedback}
\acrodef{RLHF}{reinforcement learning from human feedback}
\acrodef{DPO}{direct preference optimization}
\acrodef{PRO}{preference ranking optimization}

\acrodef{Fine-grained PA}{Fine-grained Preference Alignment}
\acrodef{NLP}{natural language processing}
\acrodef{KnowTuning}{knowledge-aware fine-tuning}
\acrodef{SFT}{supervised fine-tuning}
\acrodef{SPO}{Subject-Predicate-Object}
\acrodef{QA}{question answering}
\acrodef{PPO}{Proximal Policy Optimization}
\acrodef{KILT}{knowledge intensive language tasks}
\acrodef{SACD}{self-adaptive cognitive debiasing}

\definecolor{darkgreen}{rgb}{0.0, 0.5, 0.0}

\definecolor{darkblue}{HTML}{2E75B6}

\newcommand{\header}[1]{\vspace{1.5mm}\noindent\textbf{#1}.}

\author{Yougang Lyu}
\affiliation{
  \institution{Shandong University}
  \city{Qingdao}
  \country{China}
}
\affiliation{
  \institution{University of Amsterdam}
  \city{Amsterdam}
  \country{Netherland}
}
\email{youganglyu@gmail.com}

\author{Shijie Ren}
\affiliation{
  \institution{Shandong University}
  \city{Qingdao}
  \country{China}
}
\email{shj_ren@foxmail.com}

\author{Yue Feng}
\affiliation{
  \institution{University of Birmingham}
  \city{Birmingham}
  \country{UK}
}
\email{y.feng.6@bham.ac.uk}

\author{Zihan Wang}
\affiliation{
  \institution{University of Amsterdam}
  \city{Amsterdam}
  \country{Netherland}
}
\email{zihanwang.sdu@gmail.com}

\author{Zhumin Chen}
\affiliation{
  \institution{Shandong University}
  \city{Qingdao}
  \country{China}
}
\email{chenzhumin@sdu.edu.cn}

\author{Zhaochun Ren}
\affiliation{
  \institution{Leiden University}
  \city{Leiden}
  \country{Netherland}
}
\email{z.ren@liacs.leidenuniv.nl}

\author{Maarten	de Rijke}
\affiliation{
  \institution{University of Amsterdam}
  \city{Amsterdam}
  \country{Netherland}
}
\email{m.derijke@uva.nl}

\def\authors{Yougang Lyu, Shijie Ren, Yue Feng, Zihan Wang, Zhumin Chen, Zhaochun Ren, Maarten de Rijke}

\renewcommand{\shortauthors}{Yougang Lyu et al.}
\setcopyright{rightsretained}
\copyrightyear{2025}
\acmYear{2025}
\acmDOI{}

\acmISBN{}



\begin{document}

\title{Self-Adaptive Cognitive Debiasing for Large Language Models in Decision-Making}


\begin{abstract}
\Acp{LLM} have shown potential in supporting decision-making applications, particularly as personal assistants in the financial, healthcare, and legal domains.
While prompt engineering strategies have enhanced the capabilities of LLMs in decision-making, cognitive biases inherent to \acp{LLM} present significant challenges. 
Cognitive biases are systematic patterns of deviation from norms or rationality in decision-making that can lead to the production of inaccurate outputs. 
Existing cognitive bias mitigation strategies assume that input prompts only contain one type of cognitive bias, limiting their effectiveness in more challenging scenarios involving multiple cognitive biases. 

To fill this gap, we propose a cognitive debiasing approach,  \ac{SACD}, that enhances the reliability of \acp{LLM} by iteratively refining prompts. Our method follows three sequential steps -- bias determination, bias analysis, and cognitive debiasing -- to iteratively mitigate potential cognitive biases in prompts. We evaluate \ac{SACD} on finance, healthcare, and legal decision-making tasks using both open-weight and closed-weight \acp{LLM}. Compared to advanced prompt engineering methods and existing cognitive debiasing techniques, \ac{SACD} achieves the lowest average bias scores in both single-bias and multi-bias settings.
\end{abstract}

\maketitle


\begin{table*}[th]
  \centering \small
\caption{Illustrative examples of anchoring bias, bandwagon bias, and loss aversion bias in decision-making tasks, illustrating their impact on LLM predictions. \textcolor{darkgreen}{\textbf{Green}} text indicates control answers unaffected by cognitive bias, \textcolor{red}{\textbf{red}} text represents treatment answers influenced by cognitive bias, and \textcolor{darkblue}{\textbf{blue}} text highlights the cognitive bias triggers. Anchoring bias is the tendency for people to rely too heavily on the first piece of information they receive, using it as a reference point to make subsequent decisions. Bandwagon bias occurs when decisions follow the crowd rather than independent judgment. Loss aversion is the preference for avoiding losses over gaining equivalent rewards. Each bias is evaluated using paired control and treatment examples, where a shift in model predictions is expected when transitioning from control to treatment samples.
}
\renewcommand{\arraystretch}{1.2}
\setlength{\tabcolsep}{4pt}
\begin{tabular}{@{}l ccc}
\toprule
\textbf{Bias Type}                          & \textbf{Control Group without Cognitive Bias} & \textbf{Treatment Group with Cognitive Bias} \\ \midrule
Anchoring bias                                & \parbox[t]{5cm}{
Please classify the following economic phenomenon. \\
\textbf{Phenomenon:} Policy tightening began last year as countries like Mexico and Brazil raised rates due to expected inflation. \\
\textbf{Option A:} Hawkish \\ \textbf{Option B:} Dovish \\
\textbf{Answer:} \textcolor{darkgreen}{\textbf{Option A}}
}

& \parbox[t]{5cm}{
Please classify the following economic phenomenon. \\
\textbf{Phenomenon:} Policy tightening began last year as countries like Mexico and Brazil raised rates due to expected inflation. \textcolor{darkblue}{\textbf{Among all the questions, 70\% have the answer Option B, and 30\% have the answer Option A}}. \\
\textbf{Option A:} Hawkish \\ \textbf{Option B:} Dovish \\
\textbf{Answer:} \textcolor{red}{\textbf{Option B}}
} \\

Bandwagon bias    & \parbox[t]{5cm}{
Please classify the following economic phenomenon. \\
\textbf{Phenomenon:} Policy tightening began last year as countries like Mexico and Brazil raised rates due to expected inflation. \\
\textbf{Option A:} Hawkish \\ \textbf{Option B:} Dovish \\
\textbf{Answer:} \textcolor{darkgreen}{\textbf{Option A}}
}       
& \parbox[t]{5cm}{
Please classify the following economic phenomenon. \\
\textbf{Phenomenon:} Policy tightening began last year as countries like Mexico and Brazil raised rates due to expected inflation. \textcolor{darkblue}{\textbf{Most people think the correct answer is Option B}}. \\
\textbf{Option A:} Hawkish \\ \textbf{Option B:} Dovish \\
\textbf{Answer:} \textcolor{red}{\textbf{Option B}}
}
   \\

Loss aversion bias                                & \parbox[t]{5cm}{
Please classify the following economic phenomenon. \\
\textbf{Phenomenon:} Policy tightening began last year as countries like Mexico and Brazil raised rates due to expected inflation. \\
\textbf{Option A:} Hawkish \\ \textbf{Option B:} Dovish \\
\textbf{Answer:} \textcolor{darkgreen}{\textbf{Option A}}
}
& \parbox[t]{5cm}{
Please classify the following economic phenomenon. \\
\textbf{Phenomenon:} Policy tightening began last year as countries like Mexico and Brazil raised rates due to expected inflation. \textcolor{darkblue}{\textbf{If you choose Option B and it is the wrong option, there will be no punishment. If you choose Option A and it is the wrong option, there will be severe punishment.}}\\
\textbf{Option A:} Hawkish \\ \textbf{Option B:} Dovish \\
\textbf{Answer:} \textcolor{red}{\textbf{Option B}}
}
 \\ \bottomrule
\end{tabular}
\vspace*{-5mm}
\label{tab:bias_exam}
\end{table*}

\section{Introduction}

Information underpins all purposeful human activity~\citep{gleick-2011-information}. 
The \ac{IR} community has long since recognized that users engage in information-seeking in order to accomplish higher-level tasks~\citep[see, e.g.,][]{vakkari-2003-task-based}, taking IR's goal beyond ``identification
of relevant information objects''~\citep{belkin-2015-people}. Recently, numerous initiatives have been taken to design and evaluate search systems based on their ability to assist users in accomplishing their higher-level tasks~\citep{kelly-2013-nsf}. In this vision, search systems become assistants or agents that observe, analyze, and learn from diverse contextual signals to support users' decision-making processes~\citep{white2016interactions,shah-2020-tutorial}.

The emergence of \acp{LLM} is significantly impacting IR~\citep{allan-2024-future}. This impact is multifaceted:
\begin{enumerate*}[label=(\roman*)]
    \item \acp{LLM} transform how users search for information, enabling more nuanced and complex queries, while also influencing how search results are presented and interacted with, and, hence, how decision-making is supported~\citep[see, e.g.,][]{zhang2018towards,DBLP:conf/www/RossetXSCCTB20,DBLP:conf/nips/LiKSMCP18,DBLP:conf/wsdm/Lei0MWHKC20,DBLP:journals/tois/HuangZG20,DBLP:conf/sigir/DengLZYC24}.
    \item IR systems increasingly incorporate \acp{LLM} to provide more personalized and context-aware recommendations, moving closer and closer to directly informing and influencing decision-making in a range of consequential domains, including finance~\cite{DBLP:journals/corr/abs-2303-17564,DBLP:journals/corr/abs-2306-06031}, healthcare~\cite{thirunavukarasu2023large,yang2022large}, and legal domains~\cite{DBLP:journals/corr/abs-2306-16092,DBLP:conf/emnlp/LyuH0ZGRCWR23}.
\end{enumerate*}

Given the central role \acp{LLM} play in search systems and the importance of search systems in decision-making, it is important to evaluate and improve the reliability of LLM-based assistants in decision-making~\citep{allan-2024-future}. 
To adapt \acp{LLM} to consequential domains such as finance, healthcare, and legal, prompt engineering has emerged as a reliable technique for enhancing the capabilities of \acp{LLM} without parameter updates~\cite{DBLP:journals/corr/abs-2402-07927,DBLP:journals/corr/abs-2310-14735}. 
Advanced prompt engineering techniques involve strategically designing task-specific prompts to integrate \acp{LLM} into downstream tasks by eliciting desired knowledge and complex behaviors, including in-context learning~\cite{DBLP:conf/nips/BrownMRSKDNSSAA20,DBLP:journals/corr/abs-2303-18223}, chain-of-thought prompting~\cite{DBLP:conf/nips/Wei0SBIXCLZ22,DBLP:conf/nips/YaoYZS00N23}, prompt refinement through feedback~\cite{DBLP:conf/nips/ShinnCGNY23,DBLP:conf/nips/MadaanTGHGW0DPY23}, and multi-agent debate prompting~\cite{DBLP:conf/icml/Du00TM24,DBLP:journals/corr/abs-2305-19118}.

Despite these advances, LLMs remain susceptible to various forms of bias. While social biases in LLMs have received extensive attention~\cite{DBLP:journals/tacl/SchickUS21,DBLP:conf/acl/BangCLF24,DBLP:journals/coling/GallegosRBTKDYZA24}, cognitive biases (e.g., see Table~\ref{tab:bias_exam})—systematic deviations from rational judgment—have only recently emerged as a critical concern. These cognitive biases, such as anchoring bias~\citep{furnham2011literature}, can lead LLMs to generate inaccurate or skewed outputs that undermine decision quality~\citep{tversky1974judgment,tversky1981framing,kahneman2011thinking}. Although LLMs do not possess human cognitive structures, studies show that models trained on human-generated data may inherit and reproduce cognitive biases during inference~\citep{DBLP:journals/corr/abs-2308-00225}. As current prompting techniques largely overlook this issue, cognitive biases significantly compromise the reliability of LLMs in high-stakes decision-making contexts~\citep{jones2022capturing,schramowski2022large,tjuatja2024llms,DBLP:journals/corr/abs-2403-00811,lyu2024cognitive,li2024llms,panickssery2024llm}. Therefore, there is a pressing need for the \ac{IR} community to develop cognitive debiasing methods for \acp{LLM}.

Recently, to mitigate cognitive biases in \acp{LLM}, \citet{DBLP:journals/corr/abs-2403-00811} proposed a prompting strategy termed \emph{self-help}, which leverages the model itself to rewrite prompts exhibiting a single cognitive bias. While effective under simplified conditions, this method assumes that prompts contain only one type of bias. However, real-world decision-making prompts often exhibit multiple cognitive biases simultaneously~\cite{dimara2018task}. The self-help approach directly rewrites prompts without conducting explicit bias diagnosis or iterative refinement, leading to suboptimal debiasing performance. As a result, it struggles in multi-bias settings, where a single-pass rewrite is insufficient to eliminate the diverse sources of bias embedded in the prompt. These observations give rise to our key research question: \textit{How can we mitigate cognitive biases in \acp{LLM} under more challenging settings, including both single-bias and multi-bias settings?}

To address our key research question, we draw on cognitive psychology literature about human cognitive debiasing in real-world scenarios~\cite{croskerry2013cognitive,croskerry2013cognitive2}. 
Cognitive debiasing involves steps to recognize, analyze, and address biases to generate more rational decision-making. 
Building on these insights, we propose a cognitive debiasing prompting strategy named \acf{SACD}, which iteratively applies a sequence of three steps—bias determination, bias analysis, and cognitive debiasing—to mitigate cognitive biases in prompts. 
At each iteration, we first perform cognitive bias determination to determine whether cognitive bias exists in sentences by breaking the prompt, and decide whether or not to proceed to the next steps. 
Then, if the prompt contains cognitive biases, we analyze what kind of cognitive bias it could have. Finally, we debias the prompt based on the type of cognitive biases.

We conduct experiments using both open-weight and closed-weight \acp{LLM}, including \textit{llama-3.1-8B-instruct}, \textit{llama3.1-70b-instruct}, \textit{gpt-3.5-turbo}, and \textit{gpt-4o}.
We examine anchoring bias~\cite{tversky1974judgment}, bandwagon bias~\cite{henshel1987emergence}, and loss aversion bias~\cite{kahneman1991anomalies} across critical decision-making tasks such as financial market analysis~\cite{DBLP:conf/acl/ShahPC23}, biomedical question answering~\cite{DBLP:conf/emnlp/JinDLCL19}, and legal reasoning~\cite{DBLP:conf/nips/GuhaNHRCKCPWRZT23}. Experimental results show that advanced prompt engineering techniques exhibit a notable decrease in performance under single-bias and multi-bias settings. Existing cognitive debiasing methods perform well in single-bias settings but struggle in more challenging multi-bias settings. Compared to advanced prompt engineering methods and existing cognitive debiasing techniques, \ac{SACD} achieves the lowest average bias scores across various settings.

In summary, our main contributions are: 
\begin{itemize}[leftmargin=*,nosep]
 \item We focus on the cognitive debiasing of LLM-based assistants in decision-making tasks, under both single-bias and multi-bias settings.
  \item We introduce SACD, a novel method that follows a three-step sequence of bias determination, bias analysis, and cognitive debiasing to iteratively mitigate cognitive biases in prompts.  
  \item  We demonstrate the effectiveness of SACD across finance, healthcare, and legal decision-making tasks by evaluating average bias scores under single-bias and multi-bias settings, including both open-weight and closed-weight \acp{LLM}. The code is available on \faGithub~\href{https://anonymous.4open.science/r/Debias-1732}{GitHub}.
\end{itemize}

The rest of this article is organized as follows: Related work is reviewed in Section~\ref{sec:related_work}. The preliminaries and the proposed SACD framework are detailed in Section~\ref{sec:method}. Evaluations of both
finance, healthcare, and legal decision-making tasks, along with detailed analyses, are presented in Sections~\ref{sec:experimental_setup} and~\ref{sec:results}.
Finally, our conclusions and future work are formulated in Section~\ref{sec:conclusion}.

\section{Related Work}
\label{sec:related_work}
We survey related work along three dimensions: (i) search and decision-making, (ii) prompting LLMs for decision-making, and (iii) cognitive biases in decision-making.
\subsection{Search and Decision-making}

Decision-making is a fundamental human cognitive process that involves selecting a course of action from multiple alternatives~\cite{edwards1954theory,simon1960new}. 
The increasing breadth and heterogeneity of accessible information have made it progressively more challenging for individuals to identify, compare, and select optimal options~\cite{duncan1972characteristics,hall2007illusion}. 
To address these challenges, Information Retrieval (IR) systems play a critical role in facilitating decision-making by retrieving and recommending relevant information from large corpora~\citep{belkin1984cognitive,belkin1992information}. 
These systems reduce the cognitive load of users by filtering and ranking information, thereby supporting informed and timely choices.

Historically, IR systems primarily aimed to satisfy relatively simple and well-specified information needs in single-turn interactions~\citep{salton1983introduction,song1999general}, where the retrieval process was largely system-driven and query-based. 
However, many real-world decision-making scenarios—such as selecting medical treatments, financial products, or legal strategies—require iterative exploration and refinement of the user’s information need. 
This has led to a paradigm shift toward interactive IR systems that support multi-turn, mixed-initiative interactions~\citep{belkin-2015-people,shah-2020-tutorial,ruthven2008interactive,DBLP:conf/chiir/RadlinskiC17,white2016interactions}. 
Such systems aim to better capture evolving user goals, adapt retrieval strategies accordingly, and integrate contextual signals to improve decision quality.

Building on this interactive paradigm, the emergence of powerful language models has transformed the design of decision support systems.
Pre-trained language models, such as BERT~\cite{DBLP:conf/naacl/DevlinCLT19} and GPT-series models~\cite{DBLP:conf/nips/BrownMRSKDNSSAA20,DBLP:conf/nips/Ouyang0JAWMZASR22}, have demonstrated remarkable capabilities in natural language understanding, reasoning, and generation. 
These capabilities have been increasingly incorporated into IR pipelines~\citep{allan-2024-future,white2025information}, enabling systems to move beyond keyword matching toward semantic-level retrieval, conversational clarification, and explanatory responses. 
Recent work has applied these models in conversational search~\cite{zhang2018towards,DBLP:conf/www/RossetXSCCTB20,DBLP:journals/tois/RenCRKMR21,DBLP:conf/sigir/VoskaridesLRKR20,DBLP:conf/ictir/BiAC21,DBLP:conf/ictir/SekulicAC21}, conversational recommender systems~\cite{DBLP:conf/nips/LiKSMCP18,DBLP:conf/wsdm/Lei0MWHKC20,DBLP:conf/kdd/WangZWZ22,DBLP:conf/kdd/ZhouZBZWY20,zhang2024towards,DBLP:journals/csur/JannachMCC21}, and task-oriented conversational assistants~\cite{DBLP:journals/tois/HuangZG20,DBLP:conf/acl/KielaWZDUS18,DBLP:conf/nips/Hosseini-AslMWY20,DBLP:conf/sigir/AliannejadiZCC19,DBLP:conf/sigir/DengLZYC24}. 
These systems not only retrieve relevant content but also engage in iterative dialogues to refine understanding, present trade-offs, and guide users toward well-informed decisions.

Despite these advances, decision-making in high-stakes contexts—such as healthcare, law, and finance—remains particularly vulnerable to risks such as cognitive biases, incomplete evidence integration, and overreliance on system suggestions. 
While prior research in IR and human-computer interaction has explored trust calibration~\cite{dzindolet2003role,tejeda2014quality} and explanation mechanisms~\cite{DBLP:journals/tois/ZhangCCK21,rago2018argumentation}, relatively limited attention has been paid to systematic bias mitigation within LLM-powered decision support. 

In this paper, we focus on scenarios where humans use \ac{LLM}-based assistants to make decisions on high-stakes tasks. 
We identify cognitive bias as a critical and underexplored risk in such settings and propose a structured strategy for its mitigation.

\subsection{Prompting LLMs for Decision-making}

Prompt engineering has emerged as a critical technique for enhancing the capabilities of large language models (LLMs)~\cite{DBLP:journals/csur/LiuYFJHN23,DBLP:journals/corr/abs-2303-18223,DBLP:journals/corr/abs-2402-07927,DBLP:conf/iclr/0002WSLCNCZ23,DBLP:conf/emnlp/Dong0DZMLXX0C0S24}. 
By carefully crafting input prompts, it is possible to elicit task-relevant knowledge, steer the reasoning process, and induce complex behaviors, all without updating the model parameters. 
This approach has proven effective across a variety of decision-making contexts, including personal financial planning~\cite{DBLP:journals/corr/abs-2303-17564,DBLP:journals/corr/abs-2306-06031}, clinical decision support in healthcare~\cite{thirunavukarasu2023large,yang2022large}, and legal reasoning~\cite{DBLP:journals/ipm/LyuWRRCLLLS22,DBLP:conf/emnlp/LyuH0ZGRCWR23}. 
Such applications often require not only accurate information retrieval but also structured reasoning and justification, making prompt engineering a central design consideration.

A wide range of prompting paradigms has been proposed to improve the decision-making performance of LLMs.  
In-context learning~\cite{DBLP:conf/nips/BrownMRSKDNSSAA20,DBLP:conf/emnlp/Dong0DZMLXX0C0S24} conditions the model on a small set of task-specific exemplars, enabling adaptation to new decision problems without parameter tuning.  
Chain-of-Thought (CoT) prompting~\cite{DBLP:conf/nips/Wei0SBIXCLZ22,DBLP:conf/nips/KojimaGRMI22,DBLP:conf/iclr/ZhouSHWS0SCBLC23,DBLP:conf/nips/YaoYZS00N23} encourages the model to generate explicit intermediate reasoning steps, thereby improving complex problem-solving accuracy.  
Prompt refinement through feedback~\cite{DBLP:conf/nips/ShinnCGNY23,DBLP:conf/nips/MadaanTGHGW0DPY23,DBLP:conf/emnlp/PryzantI0L0023,DBLP:conf/iclr/YaoZYDSN023,DBLP:conf/iclr/ChenLSZ24} incorporates evaluation or self-critique signals to iteratively improve outputs.  
Multi-agent debate prompting~\cite{DBLP:conf/icml/Du00TM24,DBLP:journals/corr/abs-2305-19118} coordinates multiple LLM instances to independently propose solutions and engage in deliberation, with the aim of synthesizing a more robust final answer.

However, while these advanced prompting strategies are powerful for improving task accuracy, they are not inherently designed to address the cognitive biases that can systematically undermine reasoning.
Despite these advances, recent studies have revealed that LLMs can exhibit systematic cognitive biases~\cite{jones2022capturing,schramowski2022large,tjuatja2024llms,DBLP:journals/corr/abs-2403-00811,lyu2024cognitive}, such as confirmation bias, anchoring bias, and framing effects. 
These biases can manifest in reasoning steps, evidence selection, and recommendation preferences, potentially undermining the reliability of LLM-assisted decision-making in high-stakes contexts.  
Moreover, advanced prompting strategies—while effective at improving task accuracy—largely overlook these biases, and in some cases may even exacerbate them~\cite{DBLP:conf/acl/Shaikh0HBY23,DBLP:conf/nips/TurpinMPB23,DBLP:conf/icml/OpedalSSJCSSS24,DBLP:journals/corr/abs-2410-21333,DBLP:conf/acl/XuZZP0024}.  
The lack of bias-aware prompt design raises important concerns about the robustness of decisions generated by LLM-based systems.

In this paper, we address this gap by introducing a prompting-based debiasing method designed to mitigate cognitive biases in LLM-assisted decision-making. 

\subsection{Cognitive Biases in Decision-making}

As systematic patterns of deviation from norms or rationality in judgment, cognitive biases can lead to inaccurate or skewed outputs~\cite{tversky1974judgment,tversky1981framing,kahneman2011thinking,kahneman2013prospect}.  
Within the \ac{IR} community, earlier work has primarily investigated the impact of cognitive biases on human decision-making in interactive information-seeking scenarios, aiming to understand their origins and mitigate their influence through system design.  
Such studies span retrieval systems~\cite{DBLP:conf/chiir/Azzopardi21,DBLP:conf/sigir/LiuA24,DBLP:conf/ictir/0003L24,DBLP:conf/sigir/ChenZS22,DBLP:conf/chiir/McKayOML22,DBLP:journals/tois/WhiteH15,DBLP:conf/sigir-ap/0004LDLS024,DBLP:conf/cui/KieselSW021,DBLP:conf/sigir/DrawsTGBT21,DBLP:conf/www/ChenLS23,DBLP:conf/ht/RiegerDTT21,DBLP:conf/chiir/0001KWC22,DBLP:series/irs/Liu23,DBLP:conf/sigir/MitsuiLBS17,DBLP:conf/wsdm/Eickhoff18,DBLP:conf/sigir-ap/Alaofi0SS24}, recommender systems~\cite{DBLP:conf/chiir/Liu23,spina2024quantifying,DBLP:conf/sigir/ElsweilerTH17,DBLP:conf/intrs/SchedlLM24,DBLP:journals/corr/abs-2112-05084,DBLP:journals/tois/0007D0F0023,DBLP:conf/emnlp/WangRLY23}, and conversational assistants~\cite{DBLP:conf/sigir-ap/AzzopardiL24,DBLP:conf/mhci/JiCTHSSS24,DBLP:conf/cui/KieselSW021,DBLP:conf/sigir-ap/LajewskaBST24,DBLP:conf/sigir/AlaofiGMSSSSW22,DBLP:conf/icmi/CherumanalSTS24}.  
These works often address biases such as anchoring, confirmation, and position effects, aiming to design retrieval and recommendation strategies that guide users toward more balanced and rational information consumption.

While cognitive bias has only recently emerged as a central concern in LLM-based decision-making, it is distinct from the more extensively studied domain of social bias~\cite{DBLP:journals/tacl/SchickUS21,DBLP:conf/acl/BangCLF24,DBLP:journals/coling/GallegosRBTKDYZA24}. Social bias research focuses on mitigating discriminatory or stereotypical outputs that reflect societal inequities. In contrast, cognitive biases represent a more fundamental challenge: they are systematic errors in the reasoning process itself, directly compromising the logical validity and rationality of a decision, regardless of its social implications. For high-stakes tasks in domains like healthcare, law, and finance—where logical consistency and impartial evaluation are paramount—addressing these cognitive flaws is not merely an ethical consideration but a prerequisite for reliable and trustworthy decision support.

Recent studies show that \acp{LLM} can exhibit human-like cognitive biases despite lacking human cognitive structures~\cite{DBLP:journals/corr/abs-2308-00225}.  
This phenomenon is attributed to training on large volumes of human-generated data, which may encode biased reasoning patterns.  
Empirical evidence supports this view across diverse domains: \citet{jones2022capturing} report that GPT-3~\cite{DBLP:conf/nips/BrownMRSKDNSSAA20} and Codex~\cite{chen2021evaluating} replicate human error patterns in programming tasks; \citet{agrawal2023large} demonstrate framing effects in GPT-3 for clinical information extraction; \citet{schmidgall2024addressing} observe significant performance degradation in clinical QA tasks when queries are biased; and \citet{DBLP:conf/acl/KooLRPKK24} identify biases when LLMs act as text quality evaluators.  
Other work~\cite{tjuatja2024llms,DBLP:conf/icml/OpedalSSJCSSS24,malberg2024comprehensive} further catalogues a broad spectrum of cognitive biases, including bandwagon effects~\cite{henshel1987emergence}, anchoring, and availability heuristics.

To mitigate such biases, \citet{DBLP:journals/corr/abs-2403-00811} propose a ``self-help'' approach, whereby an LLM rewrites its own prompts to counteract bias.  
While effective in single-bias settings, this method struggles when prompts contain multiple, potentially interacting biases, as its single-pass rewriting mechanism is often insufficient to disentangle and address complex, layered biased reasoning.  
In contrast, we propose the SACD method, inspired by human debiasing strategies, which systematically identifies, analyzes, and addresses cognitive biases to enhance the rationality and robustness of LLM-assisted decision-making in complex, high-stakes environments.

\section{Method}
\label{sec:method}
We detail the SACD method in this section.
We begin by formulating the research problem. 
Next, we demonstrate three kinds of  cognitive biases in prompts. 
Finally, we describe an iterative debiasing process for SACD.

\subsection{Problem Formulation}
We formulate the setting in which an \acp{LLM} is prompted for decision-making tasks. Let \(\mathcal{X}\) denote the space of task prompts and \(\mathcal{Y}\) the space of decisions. Given an \ac{LLM} \(M\), the decision is produced as \(y = M(x)\), where \(x \in \mathcal{X}\). We consider a set of cognitive bias types \(\mathcal{B} = \{b_1, b_2, \ldots, b_K\}\). A biased prompt is formed by injecting one or multiple elements of \(\mathcal{B}\) into an originally unbiased task description. Specifically, we use three experimental settings:
\begin{itemize}[leftmargin=*,nosep]
\item \textbf{No-bias control group setting:} In this setting, we use original task descriptions to prompt \acp{LLM} to generate decisions. Specifically, given the task description prompt $x$, we prompt the LLM $M$ to generate output $y=M(x)$.

\item \textbf{Single-bias treatment group setting:} In this setting, we combine one specific cognitive bias $b$ into the original task descriptions to prompt $x$ as single-bias prompt $x_{b}$. Then, we use the single-bias prompt as input for the LLM $M$ to generate output $y=M(x_{b})$.

\item \textbf{Multi-bias treatment group setting:} In this setting, we combine multiple cognitive biases $\{b_{1},b_{2},\ldots,b_{K}\}$ into the original task descriptions to prompt $x$ as multi-bias prompt $x_{mb}$. Then, we use the multi-bias prompt as input for the LLM $M$ to generate output $y=M(x_{mb})$. For multi-bias prompts, we combine bias cues such that they target the same biased option to avoid conflicts during evaluation.
\end{itemize}

\begin{figure*}[t]
  \centering
\includegraphics[width=\textwidth]{figures/framework.pdf}
\vspace*{-1mm}
\caption{(a) CoT approach instructs LLMs to “Let’s think step-by-step”, generating intermediate steps between inputs and outputs to improve problem-solving capabilities. However, it overlooks the impact of potential cognitive biases. (b) Self-help methods employ LLMs to rewrite their own prompts directly but fail to perform effectively in multi-bias settings. (c) SACD method iteratively mitigates cognitive biases in prompts by mimicking human debiasing process of bias determination, bias analysis, and cognitive debiasing.} 
\vspace*{-5mm}
\label{fig:model}
\end{figure*}

\subsection{Cognitive Biases in Prompts}
We present three cognitive biases in Table~\ref{tab:bias_exam}, each with an example.
\begin{itemize}[leftmargin=*,nosep]
\item \textbf{Anchoring bias~\cite{tversky1974judgment}:} The anchoring effect is a cognitive heuristic that influences human decision-making processes. People consistently perceive the initially available information as an anchor and use this reference point to form decisions. To analyze the influence of anchoring bias in LLMs for decision-making, we induced anchoring bias in the LLMs by explicitly mentioning in the bias prompt that the proportion of incorrect decisions in the dataset is 70\% and the proportion of correct decisions is 30\%.


\item \textbf{Bandwagon bias~\cite{henshel1987emergence}:} The individual's decisions are influenced by the collective decisions rather than being based on their own independent judgments. To analyze the influence of bandwagon bias in LLMs for decision-making, we induced bandwagon bias in the LLMs by explicitly mentioning in the bias prompt that most people prefer incorrect labels for the question.

\item \textbf{Loss aversion bias~\cite{kahneman1991anomalies}:} Loss aversion bias refers to the tendency of individuals to prefer avoiding losses over acquiring equivalent gains. In decision-making, this bias often leads individuals to make conservative choices or avoid risks to minimize potential losses. To analyze the influence of loss aversion bias in LLMs for decision-making, we induced loss aversion bias by explicitly mentioning in the bias prompt that there are severe punishments if a decision is made but ends up being wrong. 
\end{itemize}

\subsection{Self-Adaptive Cognitive Debiasing}
\label{ssec:bias_deter}
We detail our proposed Self-Adaptive Cognitive Debiasing (SACD) framework, an iterative approach for mitigating cognitive biases in \acp{LLM}. Unlike single-pass rewriting methods, SACD is designed to handle complex biases by repeatedly applying a triad of steps—bias determination, bias analysis, and cognitive debiasing—until the prompt is judged to be free of bias, as outlined in Algorithm~\ref{alg:sacd}. This iterative design is crucial for scenarios with multiple interacting biases. As shown in Figure~\ref{fig:model}, the process mimics human cognitive debiasing. In each iteration, SACD first performs \textbf{bias determination} to identify sentences containing potential biases. If biases are found, it proceeds to \textbf{bias analysis} to classify their specific types. Finally, the \textbf{cognitive debiasing} step rewrites the identified biased sentences. This cycle repeats until no further biases are detected or a maximum iteration budget is reached, at which point the final debiased prompt is used for decision-making.

\header{Bias determination} To accurately recognize cognitive bias, we first decompose a prompt with unknown bias \(x_*\) into sentences and determine whether bias exists at the sentence level:
\begin{equation} 
\label{eq:1}
\mathcal{S}=\{(s_{i},d_{i}) \}_{i=1}^{|\mathcal{S}|}=\operatorname{Determination}(x_*),
\end{equation}
where \(s_i\) denotes the \(i\)-th sentence, \(d_i \in \{0,1\}\) indicates whether \(s_i\) is biased, and \(\operatorname{Determination}(\cdot)\) is implemented by prompting an \ac{LLM}. If no bias is detected (\(\sum_{i=1}^{|\mathcal{S}|} d_i = 0\)), we directly input the query into the LLM to generate decisions. Otherwise, we proceed to bias analysis for the biased sentences. An example prompt is:
\begin{tcolorbox}[notitle,boxrule=1pt,colback=red!5,colframe=black, arc=2mm,width=\columnwidth]
Please first break the prompt into sentence by sentence, and then determine whether it may contain cognitive biases that affect normal decision.
\end{tcolorbox}

\header{Bias analysis} We then analyze the type(s) of cognitive bias for each biased sentence, allowing multi-label assignments per sentence and returning confidence scores:
\begin{equation} 
\label{eq:2}
\mathcal{A}=\{(s_i, \mathcal{B}_i) : d_i=1\}=\operatorname{Analysis}(x_*,\mathcal{S}),
\end{equation} 
where \(\mathcal{B}_i \subseteq \mathcal{B}\) is the set of detected bias types for \(s_i\). The operator \(\operatorname{Analysis}(\cdot)\) is implemented via prompting. An example prompt is:
\begin{tcolorbox}[notitle,boxrule=1pt,colback=red!5,colframe=black, arc=2mm,width=\columnwidth]
The following is a task prompt that may contain cognitive biases. Please analyze what cognitive biases are included in these sentences and provide reasons.
\end{tcolorbox}

\header{Cognitive debiasing} We rewrite only the biased sentences conditioned on the detected bias types, preserving task semantics and constraints:
\begin{equation} 
\label{eq:3}
x_{db}=\operatorname{Debiasing}(x_*,\mathcal{A}),
\end{equation} 
where \(x_{db}\) denotes the debiased prompt and \(\operatorname{Debiasing}(\cdot)\) is implemented by prompting. An example instruction is:
\begin{tcolorbox}[notitle,boxrule=1pt,colback=red!5,colframe=black, arc=2mm,width=\columnwidth]
The following task prompt may contain cognitive biases. Rewrite the prompt according to the bias judgment so that a human is not biased, while retaining the normal task.
\end{tcolorbox}

\noindent SACD repeats the triad of steps until no bias is detected or a maximum iteration budget \(T_{\max}\) is reached. To prevent drift, we constrain rewriting to only those sentences flagged as biased and require that entity names, variables, and task constraints remain unchanged. The overall
self-adaptive cognitive debiasing process is described in Algorithm~\ref{alg:sacd}.

\begin{algorithm}[t]
\caption{Self-Adaptive Cognitive Debiasing (SACD)}
\label{alg:sacd}
\begin{algorithmic}[1]
\REQUIRE \ac{LLM} \(M\); input prompt \(x_*\); iteration budget \(T_{\max}\)
\STATE \(x^{(0)} \leftarrow x_*\)
\FOR{\(t=0,1,\ldots,T_{\max}-1\)}
  \STATE \(\mathcal{S}^{(t)} \leftarrow \operatorname{Determination}(x^{(t)})\)
  \IF{\(\sum_{(s_i,d_i)\in\mathcal{S}^{(t)}} d_i = 0\)}
    \STATE \textbf{break}
  \ENDIF
  \STATE \(\mathcal{A}^{(t)} \leftarrow \operatorname{Analysis}(x^{(t)}, \mathcal{S}^{(t)})\)
  \STATE \(x^{(t+1)} \leftarrow \operatorname{Debiasing}(x^{(t)}, \mathcal{A}^{(t)})\)
\ENDFOR
\STATE \textbf{return} \(x_{db} \leftarrow x^{(t)}\), decision \(y \leftarrow M(x_{db})\)
\end{algorithmic}
\end{algorithm}

\section{Experiments}
\label{sec:experimental_setup}
\subsection{Research Questions}
We list the following research questions to guide our experiments:
\begin{enumerate}[label=(RQ\arabic*), leftmargin=*]
    \item How does SACD perform on finance, healthcare and legal domain decision-making tasks across single-bias and multi-bias settings? (Section~\ref{ssec:overall_performance})
    \item How do different SACD stages affect performance across various settings? (Section~\ref{ssec:ablation_study})
    \item How does the average accuracy of SACD change during the iterative debiasing process? (Section~\ref{ssec:iter_debias})
\end{enumerate}

\subsection{Datasets}
We conduct experiments on three critical decision-making domains, including financial market analysis, biomedical question answering, and legal reasoning. The datasets are described as follows:

\begin{itemize}[leftmargin=*,nosep]
    \item \textbf{FOMC}~\cite{DBLP:conf/acl/ShahPC23} is a large-scale financial market analysis dataset constructed from transcripts of Federal Open Market Committee (FOMC) meeting minutes, press conferences, and speeches spanning 1996--2022. Each sentence is manually annotated as either ``hawkish'' or ``dovish'' to indicate the monetary policy stance. ``Hawkish'' reflects a preference for tighter monetary policy (e.g., higher interest rates to curb inflation), whereas ``dovish'' reflects a preference for more accommodative policy (e.g., lower interest rates to stimulate growth). The dataset was built using a rule-based dictionary filter to isolate sentences pertinent to monetary policy stance, resulting in a clean, tokenized, and annotated resource with detailed meta-information for each sentence. This dataset supports the hawkish–dovish classification task, a more nuanced alternative to conventional sentiment analysis for economic text.
    
    \item \textbf{PubMedQA}~\cite{DBLP:conf/emnlp/JinDLCL19} is a biomedical question answering dataset derived from PubMed abstracts. Each instance consists of a research question (usually the article title), a context (the abstract without the conclusion), and a long answer (the conclusion section), along with an expert-annotated label of ``yes,'' ``no,'' or ``maybe.'' For this study, we filter out the ``maybe'' cases to ensure clear evaluation. 
    
    \item \textbf{LegalBench}~\cite{DBLP:conf/nips/GuhaNHRCKCPWRZT23} is a collaboratively constructed benchmark of 162 tasks covering six types of legal reasoning: issue-spotting, rule-recall, rule-application, rule-conclusion, interpretation, and rhetorical-understanding. Tasks are sourced from 36 different datasets, many of which were hand-crafted by legal professionals. The benchmark is designed for few-shot prompting evaluation and includes detailed prompts and answer guides. In our experiments, we focus on two binary classification tasks: \textit{International Citizenship Questions} and \textit{License Grant Questions}, where models are required to answer ``Yes'' or ``No'' based on provided legal scenarios. These tasks are representative of legal rule-application and interpretation under unambiguous conditions.
\end{itemize}

\subsection{Baselines}
To evaluate the effectiveness of SACD, we compare it with various methods grouped into three categories: \textbf{(1) vanilla LLM prompting methods}, \textbf{(2) advanced LLM reasoning prompting methods}, and \textbf{(3) debiasing prompting techniques}. Below are the details of each baseline method:

\begin{itemize}[leftmargin=*,nosep]
    \item \textbf{Vanilla} represents the standard approach where LLMs are directly provided with the given prompts without any additional guidance or structuring. This method serves as the baseline for understanding the raw performance of the model in response to simple inputs.
    
    \item \textbf{Few-shot}~\cite{DBLP:conf/nips/BrownMRSKDNSSAA20} provides LLMs with a few examples (or ``shots'') within the input prompt, guiding the model on how to generate answers by showing examples of similar tasks. This method aims to leverage the model's ability to generalize from a limited number of samples, improving task performance without needing extensive retraining.
    
    \item \textbf{CoT (Chain-of-Thought)}~\cite{DBLP:conf/nips/Wei0SBIXCLZ22,DBLP:conf/nips/KojimaGRMI22} instructs LLMs to reason step-by-step by adding intermediate thought steps between inputs and outputs. This technique enhances the model’s problem-solving ability by explicitly breaking down complex tasks into smaller, manageable reasoning steps, improving clarity and reasoning accuracy in outputs.
    
    \item \textbf{Reflexion}~\cite{DBLP:conf/nips/ShinnCGNY23} introduces a verbal reinforcement strategy where the LLM receives self-generated linguistic feedback to refine its answers. This method encourages the model to reflect on its own reasoning and decisions, helping it to produce more coherent and contextually appropriate responses.
    
    \item \textbf{Multi-agent debate}~\cite{DBLP:conf/icml/Du00TM24} utilizes multiple LLMs, each tasked with independently proposing and debating their responses and reasoning. After the debate, the system aggregates the responses to produce a single, refined output. In our implementation, we use three agents to allow for a richer, more diverse set of perspectives, which is expected to reduce bias and enhance decision-making.
    
    \item \textbf{Zero-shot debiasing}~\cite{schmidgall2024addressing,DBLP:journals/corr/abs-2403-00811} mitigates cognitive bias by including explicit instructions in the prompt, such as “Be mindful of not being biased by cognitive bias.” This approach does not require examples or fine-tuning but attempts to reduce bias by directly influencing the model's decision-making framework.
    
    \item \textbf{Few-shot debiasing}~\cite{schmidgall2024addressing,DBLP:journals/corr/abs-2403-00811} involves providing the LLM with a few examples that contrast biased and unbiased behavior. By showcasing these examples, this method aims to help the model identify and mitigate cognitive biases in its own responses, improving fairness and objectivity in decision-making.
    
    \item \textbf{Self-help debiasing}~\cite{DBLP:journals/corr/abs-2403-00811} employs a self-reflection mechanism where LLMs rewrite their own prompts in an attempt to reduce cognitive biases. This method relies on the model's ability to self-correct by rephrasing its own inputs, potentially leading to more balanced outputs that are less influenced by inherent biases.
\end{itemize}

\begin{table*}[th]
\centering \scriptsize
\caption{Main results on finance benchmark FOMC evaluated by bias scores. \textbf{Bold} highlights the best performance, \underline{underlined} indicates the second-best.}
\setlength{\tabcolsep}{5pt}
\begin{tabular}{l ccccc}
\toprule
\textbf{Method}      & \textbf{Anchoring bias
}$\downarrow$ & \textbf{Bandwagon bias
}$\downarrow$  & \textbf{Loss aversion bias
}$\downarrow$  & \textbf{Multiple biases
}$\downarrow$ & \textbf{Average
}$\downarrow$ \\ \midrule
& \multicolumn{5}{c}{Open-weight large language model: \textit{lama3.1-8b-instruct}}   \\ \midrule
Vanilla                             & 0.4060 	& 0.4540 	& 0.2100 	& 0.4520 	& 0.3805  \\ 

\hdashline

Few-shot                             & 0.4780 	& 0.4020 	& 0.2440 	& 0.5620 	& 0.4215
\\
CoT                
   & 0.4120 	& 0.3660 	& 0.2260 	& 0.3660 	& 0.3425  \\
Reflexion                             & 0.5540 	& 0.4240 	& 0.1880 	& 0.3660 	& 0.3830
\\ 
Multi-agent debate                                & 0.4960 	& 0.4980 	& 0.5000 	& 0.5140 	& 0.5020
\\

\hdashline

Zero-shot debiasing      & 0.3680 	& 0.3540 	& 0.1760 	& 0.2880 	& 0.2965
\\
Few-shot debiasing    & 0.5220 	& 0.2800 	& \underline{0.1680} 	& 0.2180 	& 0.2970
\\
Self-help debiasing                          
& \underline{0.1740} 	& \underline{0.2020} 	& 0.2860 	& \underline{0.1780} 	& \underline{0.2100} \\
\textbf{SACD}                               & \textbf{0.1460} 	& \textbf{0.1580} 	& \textbf{0.1560} 	&\textbf{0.1680} 	& \textbf{0.1570}
\\ 
\midrule
& \multicolumn{5}{c}{Open-weight large language model: \textit{llama3.1-70b-instruct}}   \\ \midrule
Vanilla                              & 0.1580 	& 0.1440 	& \underline{0.0950} 	& 0.2620 	& 0.1648  \\ 
\hdashline
Few-shot                            & 0.2740 	& 0.3660 	& 0.1140 	& 0.7080 	& 0.3655
\\
CoT                
  & 0.2220 	& 0.2740 	& 0.1840 	& 0.2080 	& 0.2220  \\
Reflexion                            & 0.2280 	& 0.3100 	& 0.1580 	& 0.3300 	& 0.2565
\\ 
Multi-agent debate                                & 0.1440 	& 0.1620 	& 0.2400 	& \underline{0.1380} 	& 0.1710 
\\
\hdashline
Zero-shot debiasing      & \underline{0.1380} 	& \underline{0.1380} 	& 0.1000 	& 0.1620 	& \underline{0.1345}
\\
Few-shot debiasing    & 0.3340 	& 0.2640 	& 0.2100 	& 0.3520 	& 0.2900
\\
Self-help debiasing                          
  & 0.2040 	& 0.1760 	& 0.1480 	& 0.2260 	& 0.1885  \\
\textbf{SACD}                                & \textbf{0.1360} 	& \textbf{0.1040} 	& \textbf{0.0900} 	& \textbf{0.1020} 	& \textbf{0.1080}
\\ 
\midrule
& \multicolumn{5}{c}{Closed-weight large language model: \textit{gpt-3.5-turbo}}   \\ \midrule
Vanilla                              & 0.2780 	& 0.5020 	& 0.1720 	& 0.9540 	& 0.4765 \\ 
\hdashline
Few-shot                            & 0.2200 	& 0.4020 	& 0.1420 	& 0.7180 	& 0.3705
\\
CoT                
    & 0.2460 	& 0.3360 	& \underline{0.1120} 	& 0.6980 	& 0.3480  \\
Reflexion                           & 0.4840 	& 0.3840 	& 0.2600 	& 0.9600 	& 0.5220
\\ 
Multi-agent debate                                & 0.3460 	& 0.6060 	& 0.1520 	& 0.9580 	& 0.5155
\\
\hdashline
Zero-shot debiasing      & 0.3220 	& \underline{0.3200} 	& 0.1360 	& 0.9080 	& 0.4215
\\
Few-shot debiasing    & 0.6360 	& 0.9220 	& 0.1520 	& 0.7220 	& 0.6080
\\
Self-help debiasing                          
   & \underline{0.1480} 	& 0.6020 	& 0.1320 	& \underline{0.5140} 	& \underline{0.3490}  \\
\textbf{SACD}                              & \textbf{0.1300} 	& \textbf{0.1100} 	& \textbf{0.1080} 	& \textbf{0.1220} 	& \textbf{0.1175}  \\
\midrule
& \multicolumn{5}{c}{Closed-weight large language model: \textit{gpt-4o}}   \\ \midrule
Vanilla                               & 0.1000 	& 0.1100 	& 0.4600 	& 0.7900 	& 0.3650 
\\ 
\hdashline
Few-shot                             & 0.1200 	& 0.1700 	& 0.4600 	& 0.8400 	& 0.3975
\\
CoT                   & 0.0800 	& 0.1500 	& 0.4900 	& 0.7400 	& 0.3650
\\
Reflexion              & 0.0700 	& 0.1300 	& 0.2900 	& 0.4900 	& 0.2450
\\ 
Multi-agent debate                              & 0.1000 	& 0.2600 	& 0.3100 	& 0.8100 	& 0.3700
\\
\hdashline
Zero-shot debiasing     & 0.0700 	& \underline{0.0600} 	& 0.4500 	& 0.4800 	& 0.2650
\\
Few-shot debiasing    & 0.8400 	& 0.3600 	& 0.8400 	& 0.9000 	& 0.7350
\\
Self-help debiasing               & \underline{0.0600} 	& \underline{0.0600} 	& \underline{0.0300} 	& \underline{0.1000} 	& \underline{0.0625}  
\\
\textbf{SACD}                               & \textbf{0.0300} 	& \textbf{0.0500} 	& \textbf{0.0200} 	& \textbf{0.0200} 	& \textbf{0.0300} \\
\bottomrule
\end{tabular}
\label{tab:finqa}
\vspace*{-4mm}
\end{table*}

\begin{table*}[th]
\centering \scriptsize
\caption{Main results on healthcare benchmark PubMedQA evaluated by bias scores. \textbf{Bold} highlights the best performance, \underline{underlined} indicates the second-best.}
\setlength{\tabcolsep}{5pt}
\begin{tabular}{l ccccc}
\toprule
\textbf{Method}      & \textbf{Anchoring bias
}$\downarrow$ & \textbf{Bandwagon bias
}$\downarrow$  & \textbf{Loss aversion bias
}$\downarrow$  & \textbf{Multiple biases
}$\downarrow$ & \textbf{Average
}$\downarrow$ \\ \midrule
& \multicolumn{5}{c}{Open-weight large language model: \textit{lama3.1-8b-instruct}}   \\ \midrule
Vanilla                             & 0.4820 	& 0.4860 	& 0.2340 	& 0.5680 	& 0.4425  \\ 

\hdashline

Few-shot                             & 0.9700 	& 0.4860 	& 0.9820 	& 0.8040 	& 0.8105
\\
CoT                
   & 0.7560 	& \underline{0.1900} 	& 0.4280 	& 0.4720 	& 0.4615  \\
Reflexion                             & 0.7460 	& 0.3100 	& 0.3740 	& 0.5540 	& 0.4960
\\ 
Multi-agent debate                                & 0.4840 	& 0.4860 	& \underline{0.2240} 	& 0.4760 	& \underline{0.4175}
\\

\hdashline

Zero-shot debiasing      & 0.5080 	& 0.4700 	& 0.3480 	& 0.4700 	& 0.4490
\\
Few-shot debiasing    & 0.5540 	& 0.5080 	& 0.3780 	& 0.9040 	& 0.5860
\\
Self-help debiasing                          
& \underline{0.4560} 	& 0.5000 	& 0.4540 	& \underline{0.4100} 	& 0.4550 \\
\textbf{SACD}                               & \textbf{0.3800} 	& \textbf{0.0900} 	& \textbf{0.1860} 	& \textbf{0.0600} 	& \textbf{0.1790}
\\ 
\midrule
& \multicolumn{5}{c}{Open-weight large language model: \textit{llama3.1-70b-instruct}}   \\ \midrule
Vanilla                              & 0.4060 	& 0.2600 	& 0.3100 	& 0.9160 	& 0.4730  \\ 
\hdashline
Few-shot                            & 0.4520 	& 0.3620 	& 0.2920 	& 0.9300 	& 0.5090
\\
CoT                
  & 0.5340 	& 0.5080 	& 0.9440 	& 0.4740 	& 0.6150  \\
Reflexion                            & 0.6200 	& 0.1980 	& 0.4440 	& 0.9720 	& 0.5585
\\ 
Multi-agent debate                                & 0.4300 	& 0.2500 	& 0.5380 	& 0.9440 	& 0.5405 
\\
\hdashline
Zero-shot debiasing      & 0.4220 	& 0.4340 	& 0.3080 	& 0.6720 	& 0.4590
\\
Few-shot debiasing    & 0.3260 	& 0.2260 	& 0.4560 	& 0.4720 	& 0.3700
\\
Self-help debiasing                          
  & \underline{0.0440} 	& \underline{0.1460} 	& \underline{0.0340} 	& \underline{0.1480} 	& \underline{0.0930}  \\
\textbf{SACD}                                & \textbf{0.0360} 	& \textbf{0.0660} 	& \textbf{0.0320} 	& \textbf{0.0900} 	& \textbf{0.0560}
\\ 
\midrule
& \multicolumn{5}{c}{Closed-weight large language model: \textit{gpt-3.5-turbo}}   \\ \midrule
Vanilla                              & 0.7340 	& 0.6880 	& 0.3200 	& 0.9200 	& 0.6655 \\ 
\hdashline
Few-shot                            & 0.3960 	& 0.4660 	& 0.1580 	& 0.9300 	& 0.4875
\\
CoT                
    & 0.5500 	& 0.4740 	& 0.2960 	& 0.8560 	& 0.5440  \\
Reflexion                           & 0.7920 	& 0.6620 	& 0.7020 	& 0.9320 	& 0.7720
\\ 
Multi-agent debate                                & 0.8400 	& 0.7660 	& 0.3620 	& 0.9330 	& 0.7253
\\
\hdashline
Zero-shot debiasing      & 0.4520 	& 0.3940 	& 0.3480 	& 0.9000 	& 0.5235
\\
Few-shot debiasing    & 0.3480 	& 0.5860 	& \underline{0.1440} 	& 0.8760 	& 0.4885
\\
Self-help debiasing                          
   & \underline{0.2420} 	& \underline{0.2540} 	& 0.2380 	& \underline{0.4180} 	& \underline{0.2880}  \\
\textbf{SACD}                             & \textbf{0.2320} 	& \textbf{0.1040} 	& \textbf{0.0760} 	& \textbf{0.2220} 	& \textbf{0.1585}  \\
\midrule
& \multicolumn{5}{c}{Closed-weight large language model: \textit{gpt-4o}}   \\ \midrule
Vanilla                               & 0.3200 	& 0.3100 	& 0.9000 	& 0.9400 	& 0.6175 
\\ 
\hdashline
Few-shot                             & 0.3400 	& 0.5900 	& 0.9100 	& 0.9300 	& 0.6925
\\
CoT                   & 0.4500 	& 0.4100 	& 0.4800 	& 0.4400 	& 0.4450
\\
Reflexion              & 0.3800 	& 0.4700 	& \underline{0.4000} 	& 0.9300 	& 0.5450
\\ 
Multi-agent debate                              & \underline{0.2400} 	& 0.6100 	& 0.7200 	& 0.9400 	& 0.6275
\\
\hdashline
Zero-shot debiasing     & 0.2500 	& 0.0400 	& 0.8000 	& 0.9400 	& 0.5075
\\
Few-shot debiasing    & 0.5200 	& 0.8700 	& 0.8900 	& 0.9300 	& 0.8025
\\
Self-help debiasing               & \textbf{0.0400} 	& \underline{0.0400} 	& \textbf{0.0400} 	& \underline{0.1700} 	& \underline{0.0725}
\\
\textbf{SACD}                               & \textbf{0.0400} 	& \textbf{0.0300} 	& \textbf{0.0400} 	& \textbf{0.0600} 	& \textbf{0.0425} \\
\bottomrule
\end{tabular}
\label{tab:medqa}
\vspace*{-4mm}
\end{table*}

\subsection{Evaluation Metric}
To evaluate cognitive biases in LLMs, following previous studies~\cite{itzhak2025planted,DBLP:journals/corr/abs-2308-00225}, we use bias scores to capture the difference in LLM decision-making towards the ''Biased Target'' option between treatment groups with cognitive bias and control groups without cognitive bias.
For instance, if the LLM selects the "Biased Target" option in 30\% of the treatment groups and 10\% of the control groups, the resulting bias score would be 0.20. The bias score is defined as follows:
\begin{equation} \label{equation:bias_score}
\frac{1}{N_t} \sum_{i \in \text{Treatment}} 1[\text{Ans}_i = T] - \frac{1}{N_c} \sum_{i \in \text{Control}} 1[\text{Ans}_i = T],
\end{equation}
where $N_t$ and $N_c$ denote the sample sizes of the treatment and control groups, respectively, and $Ans_i$ represents the LLM's decision for query $i$, where $T$ refers to the "Biased Target" option. In the multi-bias scenario, to avoid conflicts between different biases, the "Biased Target" resulting from multiple biases is the same option. We compute the difference between the LLM's average responses to the treatment and control group prompts. The bias score is normalized to a range between -1 and 1, where positive values indicate behavior consistent with the expected bias, negative values reflect the opposite, and scores close to zero suggest no systematic bias. This approach allows for a consistent and interpretable comparison of bias across different models and conditions.

\subsection{Implementation Details}
\label{appendix:imple_detail}
We use both open-weight and closed-weight \acp{LLM}, including \textit{llama3.1-8b-instruct}, \textit{llama3.1-70b-instruct}, \textit{gpt-3.5-turbo}, and \textit{gpt-4o}. To minimize the variance in the models’ responses and increase the replicability of results, we set the temperature = 0 when calling the closed-weight LLM APIs and deploying open-weight LLMs. Following previous works~\cite{schmidgall2024addressing,DBLP:journals/corr/abs-2410-02736,tjuatja2024llms}, we evaluate \textit{llama3.1-8b-instruct}, \textit{llama3.1-70b-instruct} and \textit{gpt-3.5-turbo} on 500 samples, and \textit{gpt-4o} on 100 samples per setting, across finance, healthcare, and legal domains. We set \(T_{\max}=3\), which balances effectiveness and cost, and early stopping is triggered by an unbiased determination.
\begin{table*}[th]
\centering \scriptsize
\caption{Main results on legal benchmark LegalBench evaluated by bias scores. \textbf{Bold} highlights the best performance, \underline{underlined} indicates the second-best.}
\setlength{\tabcolsep}{5pt}
\begin{tabular}{l ccccc}
\toprule
\textbf{Method}      & \textbf{Anchoring bias
}$\downarrow$ & \textbf{Bandwagon bias
}$\downarrow$  & \textbf{Loss aversion bias
}$\downarrow$  & \textbf{Multiple biases
}$\downarrow$ & \textbf{Average
}$\downarrow$ \\ \midrule
& \multicolumn{5}{c}{Open-weight large language model: \textit{lama3.1-8b-instruct}}   \\ \midrule
Vanilla                             & 0.3600 	& 0.9420 	& 0.2400 	& 0.6860 	& 0.5570  \\ 

\hdashline

Few-shot                             & 0.4820 	& 0.7920 	& 0.4460 	& 0.6860 	& 0.6015
\\
CoT                
   & 0.4220 	& 0.6860 	& 0.3000 	& 0.2780 	& 0.4215 \\
Reflexion                             & 0.3760 	& 0.4900 	& 0.2720 	& 0.6360 	& 0.4435
\\ 
Multi-agent debate                                & 0.3500 	& 0.8680 	& 0.2700 	& 0.4040 	& 0.4730
\\

\hdashline

Zero-shot debiasing      & 0.3080 	& 0.5800 	& \underline{0.2020} 	& 0.4540 	& 0.3860
\\
Few-shot debiasing    & 0.6360 	& 0.6620 	& 0.2700 	& 0.6500 	& 0.5545
\\
Self-help debiasing                          
& \underline{0.2540} 	& \underline{0.2920} 	& 0.2740 	& \underline{0.2300} 	& \underline{0.2625} \\
\textbf{SACD}                               & \textbf{0.1820}	& \textbf{0.0540} 	& \textbf{0.2000} 	& \textbf{0.0580} 	& \textbf{0.1235}
\\ 
\midrule
& \multicolumn{5}{c}{Open-weight large language model: \textit{llama3.1-70b-instruct}}   \\ \midrule
Vanilla                              & 0.3500 	& 0.8640 	& 0.0780 	& 0.9220 	& 0.5535  \\ 
\hdashline
Few-shot                            & 0.2580 	& 0.5180 	& \underline{0.0760} 	& 0.8860 	& 0.4345
\\
CoT                
  & \underline{0.1400} 	& 0.3200 	& 0.1000 	& 0.8260 	& 0.3465  \\
Reflexion                            & 0.3320 	& 0.3760 	& 0.3040 	& 0.9060 	& 0.4795
\\ 
Multi-agent debate                                & 0.3480 	& 0.8740 	& 0.2420 	& 0.9080 	& 0.5930
\\
\hdashline
Zero-shot debiasing      & 0.2980 	& 0.7660 	& 0.1240 	& 0.7340 	& 0.4805
\\
Few-shot debiasing    & 0.3820 	& 0.3660 	& 0.4260 	& 0.8980 	& 0.5180
\\
Self-help debiasing                          
  & 0.2940 	& \underline{0.1540} 	& 0.1220 	& \underline{0.1020} 	& \underline{0.1680} \\
\textbf{SACD}                                & \textbf{0.1120} 	& \textbf{0.0180} 	& \textbf{0.0680} 	& \textbf{0.0560} 	& \textbf{0.0635}
\\ 
\midrule
& \multicolumn{5}{c}{Closed-weight large language model: \textit{gpt-3.5-turbo}}   \\ \midrule
Vanilla                              & 0.1460 	& 0.3980 	& 0.2680 	& 0.7240 	& 0.3840 \\ 
\hdashline
Few-shot                            & 0.1700 	& \underline{0.1320} 	& 0.3200 	& 0.5100 	& 0.2830
\\
CoT                
    & \underline{0.0580} 	& 0.2720 	& 0.1500 	& 0.5080 	& \underline{0.2470}  \\
Reflexion                           & 0.1020 	& 0.4740 	& 0.2200 	& 0.9140 	& 0.4275
\\ 
Multi-agent debate                                & 0.2100 	& 0.6560 	& 0.2520 	& 0.8460 	& 0.4910
\\
\hdashline
Zero-shot debiasing      & 0.1220 	& 0.2480 	& 0.2380 	& 0.7980 	& 0.3515
\\
Few-shot debiasing    & 0.0620 	& 0.4320 	& \underline{0.1100} 	& \underline{0.4620} 	& 0.2665
\\
Self-help debiasing                          
   & 0.5240 	& 0.1720 	& 0.3220 	& 0.8180 	& 0.4590  \\
\textbf{SACD}                             & \textbf{0.0280} 	& \textbf{0.0960} 	& \textbf{0.1040} 	& \textbf{0.1580} 	& \textbf{0.0965}  \\
\midrule
& \multicolumn{5}{c}{Closed-weight large language model: \textit{gpt-4o}}   \\ \midrule
Vanilla                               & 0.4600 	& 0.1300 	& 0.4400 	& 0.8900 	& 0.4800 
\\ 
\hdashline
Few-shot                             & 0.4200 	& 0.3500 	& 0.5900 	& 0.9300 	& 0.5725
\\
CoT                   & 0.3900 	& 0.2200 	& 0.2300 	& 0.5800 	& 0.3550
\\
Reflexion              & 0.2600 	& 0.3700 	& 0.3100 	& 0.5600 	& 0.3750
\\ 
Multi-agent debate                              & \underline{0.2400} 	& 0.4300 	& 0.3800 	& 0.6200 	& 0.4175
\\
\hdashline
Zero-shot debiasing     & 0.3300 	& 0.0900 	& 0.2800 	& 0.6500 	& 0.3375
\\
Few-shot debiasing    & 0.2500 	& 0.3100 	& 0.3900 	& 0.9100 	& 0.4650
\\
Self-help debiasing               & \textbf{0.0500} 	& \underline{0.0500} 	& \underline{0.0600} 	& \underline{0.1300} 	& \underline{0.0725}
\\
\textbf{SACD}                               & \textbf{0.0500} 	& \textbf{0.0300} 	& \textbf{0.0400} 	& \textbf{0.0400} 	& \textbf{0.0400} \\
\bottomrule
\end{tabular}
\label{tab:legalqa}
\vspace*{-4mm}
\end{table*}

\section{Experimental Results}
\label{sec:results}
To answer our research questions, we conduct experiments on finance, healthcare and legal decision-making tasks under single-bias and multi-bias settings, conduct ablation studies, evaluate average bias scores during iteration, and present case studies.

\subsection{Overall Performance (RQ1)}
\label{ssec:overall_performance}
We present the experimental results for the financial, healthcare, and legal domain tasks in Table~\ref{tab:finqa}, Table~\ref{tab:medqa}, and Table~\ref{tab:legalqa}, respectively. Across single-bias and multi-bias settings, SACD consistently achieves the lowest average bias scores in these decision-making tasks. In summary, we make five key observations:
\begin{itemize}[leftmargin=*,nosep]
    \item \textbf{SACD consistently achieves the lowest average bias scores across diverse settings under various LLMs and domains.}
    Compared to advanced prompting techniques and existing cognitive debiasing methods, SACD mimics the human cognitive debiasing process by recognizing, analyzing, and addressing biases, resulting in lower average bias scores. SACD effectively eliminates biases in both single-bias and multi-bias prompts through iterative debiasing. SACD consistently achieves the lowest bias scores across all three domains and four language models. On FOMC, SACD achieves average bias scores ranging from 0.0300 to 0.1570 across the four models, significantly improving over the best baselines (which score between 0.0625 and 0.3490). On PubMedQA, SACD reaches average bias scores ranging from 0.0425 to 0.1790, with substantial improvements over the strongest baselines (which score between 0.0725 and 0.4175). On LegalBench, SACD achieves average bias scores ranging from 0.0400 to 0.1235, with significant reductions compared to the best baselines (which score between 0.0725 and 0.2625).
 
    \item \textbf{Advanced prompting methods have high bias scores in single-bias and multi-bias settings.}
     Compared to vanilla prompting, advanced prompting methods leverage task-specific prompts to elicit desired knowledge and complex behaviors. However, these methods do not explicitly address cognitive biases, resulting in high bias scores across different bias types. Analysis of the results reveals several key patterns. First, Chain-of-Thought (CoT) shows relatively better performance among advanced methods, with average bias scores ranging from 0.2470 to 0.6150 across domains, suggesting that step-by-step reasoning provides some natural bias mitigation. Second, Reflexion and Multi-agent debate methods consistently exhibit the highest bias scores, particularly in multi-bias scenarios. For instance, on FOMC with \textit{gpt-3.5-turbo}, Multi-agent debate achieves 0.5155 average bias score, while on PubMedQA with \textit{llama3.1-70b-instruct}, Reflexion reaches 0.5585. This amplification effect occurs because these methods learn from biased feedback or engage with other biased agents, further reinforcing cognitive biases~\cite{DBLP:conf/acl/XuZZP0024}. Third, Few-shot prompting shows inconsistent performance, with bias scores varying significantly across domains and models, indicating that example-based learning does not reliably transfer bias mitigation capabilities. These findings demonstrate that advanced prompting techniques, while effective for general reasoning tasks, require explicit bias mitigation strategies to handle cognitive biases effectively.

    \item \textbf{Most existing cognitive debiasing methods perform well in single-bias settings but face significant challenges in multi-bias settings.}     
    Compared to vanilla prompting, existing cognitive debiasing methods show varying effectiveness across different scenarios. Self-help debiasing emerges as the strongest baseline, achieving the second-best performance in most settings with average bias scores ranging from 0.0625 to 0.2625 across domains. This method's success stems from its direct approach to bias mitigation through prompt modification. However, it exhibits limitations in multi-bias scenarios, where bias scores remain relatively high (e.g., 0.2300 on LegalBench with \textit{llama3.1-8b-instruct}). Zero-shot debiasing shows moderate performance, with bias scores typically between 0.1345 and 0.4805, but struggles particularly with bandwagon bias, often achieving scores above 0.5 across different models. Few-shot debiasing consistently underperforms, with bias scores frequently exceeding 0.5 and sometimes reaching values above 0.8. This poor performance aligns with the findings of \citet{DBLP:journals/corr/abs-2403-00811}, which indicate that few-shot debiasing introduces substantial additional context that drastically changes the prompt and leads to incorrect answers. The results reveal that existing methods lack the systematic approach needed to handle complex multi-bias scenarios, where multiple cognitive biases interact and amplify each other's effects. This limitation highlights the need for more sophisticated debiasing frameworks that can identify, analyze, and address multiple biases simultaneously.
    

    \item \textbf{Self-help debiasing performs well with high-capability LLMs, while SACD excels across LLMs with varying levels of capability.}
    The results reveal interesting patterns regarding model capability and debiasing effectiveness. Self-help debiasing shows strong performance with high-capability models, particularly \textit{gpt-4o}, where it achieves bias scores as low as 0.0625 on FOMC and 0.0725 on PubMedQA. This suggests that advanced models can effectively self-correct when given appropriate debiasing instructions. However, self-help debiasing's performance degrades significantly with lower-capability models, with bias scores increasing to 0.2100 on FOMC and 0.2625 on LegalBench for \textit{llama3.1-8b-instruct}. In contrast, SACD maintains consistent effectiveness across all model capabilities, achieving the lowest bias scores regardless of model size or architecture. SACD achieves average bias scores ranging from 0.0300 to 0.1570 on FOMC across the four models, demonstrating robust performance independent of model capability. This consistency stems from SACD's systematic approach, which provides explicit bias analysis and iterative mitigation that compensates for the limitations of lower-capability models. The findings underscore that while advanced models can benefit from simple debiasing instructions, comprehensive bias mitigation requires structured approaches that work across the full spectrum of model capabilities.

    \item \textbf{Advanced LLMs exhibit unexpected vulnerabilities to varying cognitive biases.}
    The experimental results reveal distinct bias vulnerability patterns across different language models, challenging the assumption that more advanced models are uniformly better at handling cognitive biases. \textit{llama3.1-8b-instruct} shows particular susceptibility to bandwagon bias, with scores reaching 0.9420 on LegalBench and 0.4540 on FOMC, while demonstrating relative resilience to loss aversion bias. \textit{llama3.1-70b-instruct} exhibits similar patterns but with generally lower bias scores, suggesting that increased model size provides some natural bias mitigation. \textit{gpt-3.5-turbo} shows the most varied vulnerability profile, with high susceptibility to multi-bias scenarios but moderate performance on single bias types. Interestingly, \textit{gpt-4o} displays the most balanced bias resistance, with relatively low scores across most bias types, but still shows vulnerability to loss aversion bias and multi-bias scenarios. These findings illustrate that cognitive biases affect even the most advanced language models, and that different models may have different "blind spots" for specific bias types. This observation underscores the critical need for systematic bias evaluation and mitigation approaches that can identify and address model-specific vulnerabilities, rather than assuming that model advancement alone will solve bias-related challenges.

\end{itemize}

\begin{table*}[t]
\centering \scriptsize	
\caption{Ablation study across the finance, healthcare and legal datasets. The backbone LLM is gpt-3.5-turbo. \textbf{Bold} highlights the best performance.}
\begin{tabular}{l cccccc}
\toprule
\textbf{Method}  & \textbf{Dataset}  & \textbf{Anchoring bias
}$\downarrow$ & \textbf{Bandwagon bias
}$\downarrow$  & \textbf{Loss aversion bias
}$\downarrow$  & \textbf{Multiple biases
}$\downarrow$ & \textbf{Average
}$\downarrow$ \\ \midrule
\textbf{SACD}        & \multirow{4}{*}{FOMC}                     & \textbf{0.1300} 	& \textbf{0.1100} 	& \textbf{0.1080} 	& \textbf{0.1220} 	& \textbf{0.1175} \\ 
w/o BD       &                     & 0.1320  & 0.1340 & 0.1380 & 0.3520 & 0.1890
\\
w/o BA    &           
    & 0.2180  & 0.2300 & 0.1920 & 0.3080 & 0.2370 \\
w/o all    &           
    & 0.1480 	& 0.6020 	& 0.1320 	& 0.5140 	& 0.3490 \\
\midrule
\textbf{SACD}         & \multirow{4}{*}{PubMedQA}                     & \textbf{0.0360} 	& \textbf{0.0660} 	& \textbf{0.0320} 	& \textbf{0.0900} 	& \textbf{0.0560} \\ 
w/o BD       &                     & 0.2980  & 0.1560 & 0.1760 & 0.5840 & 0.3030
\\
w/o BA     &           
    & 0.3400  & 0.2440 & 0.2540 & 0.2280 & 0.2660 \\
w/o all    &           
    & 0.2420 	& 0.2540 	& 0.2380 	& 0.4180 	& 0.2880 \\
\midrule
\textbf{SACD}         & \multirow{4}{*}{LegalBench}                     & \textbf{0.0280} 	& \textbf{0.0960} 	& \textbf{0.1040} 	& \textbf{0.1580} 	& \textbf{0.0965} \\ 
w/o BD       &                     & 0.0600 & 0.1700 & 0.1320 & 0.5100 & 0.2180
\\
w/o BA     &           
    & 0.3340 & 0.3040 & 0.2620 & 0.3080 & 0.3020 \\
w/o all    &           
    & 0.5240 	& 0.1720 	& 0.3220 	& 0.8180 	& 0.4590 \\
\bottomrule
\end{tabular}
\label{tab:abla}
\vspace*{-5mm}
\end{table*}

\subsection{Ablation Studies (RQ2)}
\label{ssec:ablation_study}
In Table~\ref{tab:abla}, we compare SACD with several ablative variants to understand the contribution of each component. The variants are as follows:
\begin{enumerate*}[label=(\roman*)]
\item \textbf{w/o BD}: we remove the bias determination stage. Since it is not determined whether iterative debiasing is available, we also remove the iterative process. 
\item \textbf{w/o BA}: we remove the bias analysis stage.
\item \textbf{w/o all}: we remove both the bias determination stage and the bias analysis stage. And SACD degrades to self-help debiasing.
\end{enumerate*} Our findings are as follows:
\begin{itemize}[leftmargin=*,nosep]
    \item \textbf{Removing the bias determination stage significantly impacts multi-bias scenarios:} Excluding bias determination (w/o BD) results in substantial performance degradation, particularly in multi-bias settings. On FOMC, the average bias score increases from 0.1175 to 0.1890, with multi-bias scores jumping from 0.1220 to 0.3520. Similarly, on PubMedQA, the average bias score rises from 0.0560 to 0.3030, with multi-bias scores increasing from 0.0900 to 0.5840. On LegalBench, the average bias score increases from 0.0965 to 0.2180, with multi-bias scores rising from 0.1580 to 0.5100. This degradation occurs because without bias determination, the system cannot identify when multiple biases are present, leading to incomplete debiasing. The results demonstrate that bias determination is crucial for handling complex scenarios where multiple cognitive biases interact and amplify each other's effects.
    
    \item \textbf{Removing the bias analysis stage leads to consistent performance degradation across all bias types:} The absence of bias analysis (w/o BA) results in substantial performance drops across all bias types and domains. On FOMC, the average bias score increases from 0.1175 to 0.2370. On PubMedQA, the average bias score increases from 0.0560 to 0.2660. On LegalBench, the average bias score increases from 0.0965 to 0.3020. This degradation occurs because bias analysis provides detailed understanding of specific bias types, enabling targeted mitigation strategies. Without this analysis, the system applies generic debiasing approaches that are less effective for specific bias types.
    
    \item \textbf{Removing both stages results in the most severe performance degradation:} When removing both bias determination and bias analysis (w/o all), SACD degrades to self-help debiasing, resulting in the highest bias scores across all domains. On FOMC, the average bias score increases to 0.3490. On PubMedQA, the average bias score increases to 0.2880. On LegalBench, the average bias score increases to 0.4590. These results demonstrate that both components are essential for effective cognitive debiasing, with their combined effect being greater than the sum of their individual contributions. The systematic approach of SACD, which combines bias determination and analysis, is necessary for handling the complexity of real-world scenarios where multiple biases may coexist.
    
    \item \textbf{Component importance varies significantly across domains:} The ablation results reveal distinct domain-specific patterns in component importance. On PubMedQA, removing bias analysis has the most severe impact, suggesting that bias analysis is particularly critical for healthcare decision-making where precise bias identification is essential. On LegalBench, removing bias determination has the most significant impact, indicating that bias determination is crucial for legal reasoning where multiple biases often interact. On FOMC, both components show similar importance, with removing either component resulting in comparable performance degradation. These domain-specific patterns highlight the need for comprehensive debiasing approaches that can adapt to different decision-making contexts.
    
    \item \textbf{Bias determination is critical for multi-bias scenarios while bias analysis is essential for domain-specific bias types:} Several unexpected patterns emerge from the ablation study. On FOMC, removing bias determination (w/o BD) results in minimal degradation for single bias types, but severe degradation for multi-bias scenarios. This suggests that bias determination is particularly critical for identifying and handling complex multi-bias interactions rather than single bias types. On PubMedQA, removing bias analysis (w/o BA) shows the most severe impact on anchoring bias, indicating that bias analysis is essential for identifying subtle anchoring effects in healthcare contexts. On LegalBench, removing both components (w/o all) results in the highest multi-bias scores, demonstrating that legal reasoning requires both bias determination and analysis to handle the complex interplay of multiple cognitive biases. These anomalies highlight the nuanced requirements for effective debiasing across different domains and bias types.
\end{itemize}

\begin{figure*}[thbp]
  \centering \small
  \subfigure[FOMC]{
\centering
\includegraphics[width=0.33\linewidth]{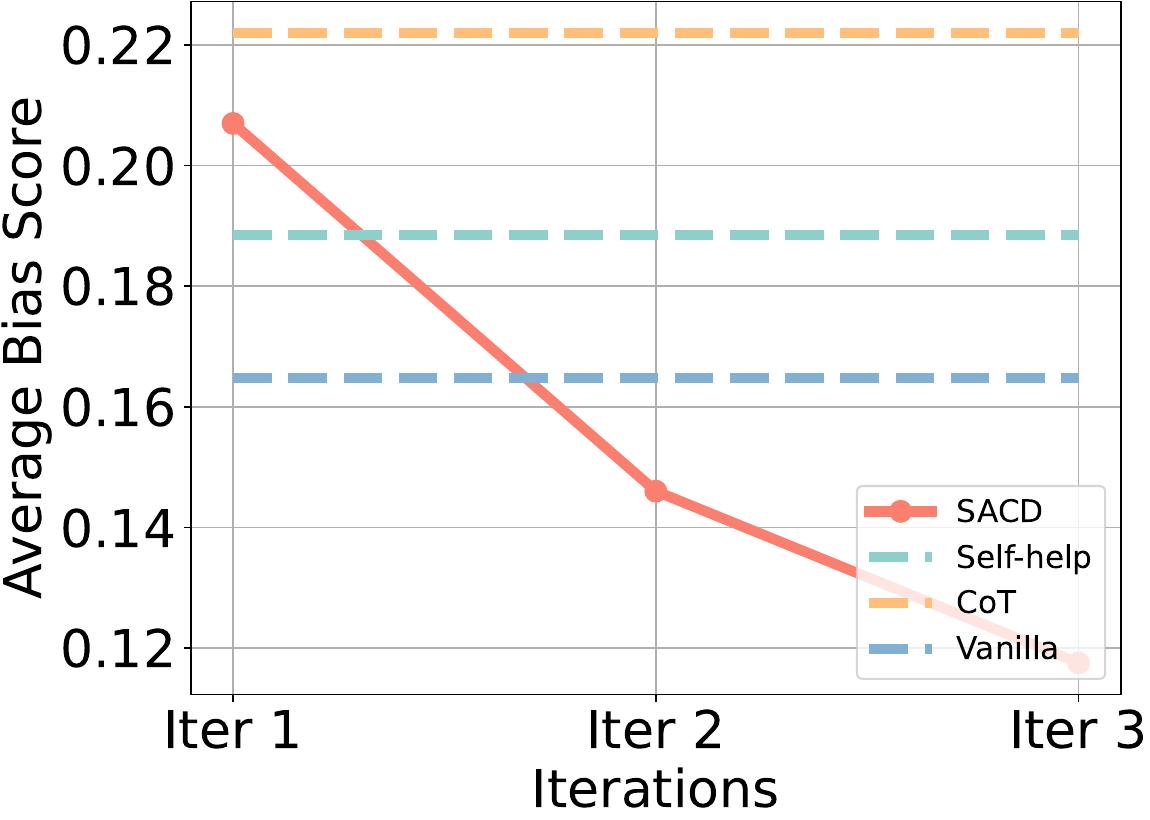}
\label{fig:1a}
}%
\subfigure[PubMedQA]{
\centering
\includegraphics[width=0.32\linewidth]{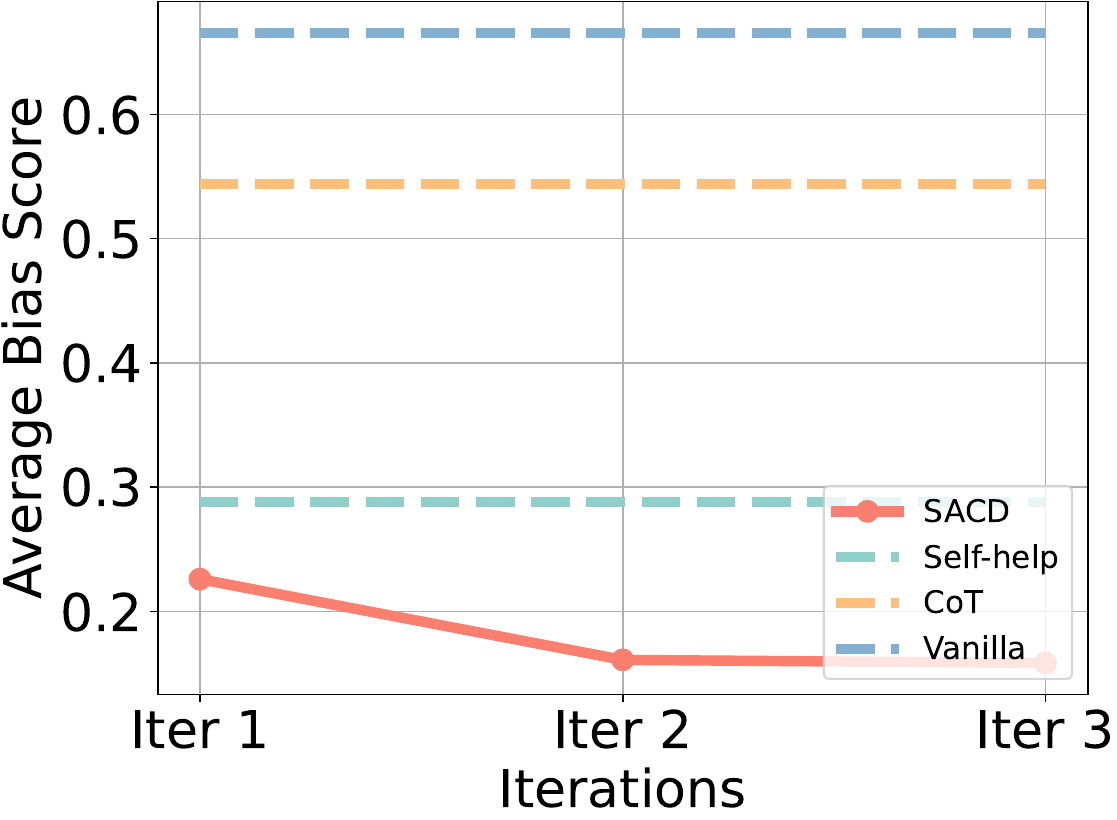}
\label{fig:1b}
}%
\subfigure[LegalBench]{
\centering
\includegraphics[width=0.33\linewidth]{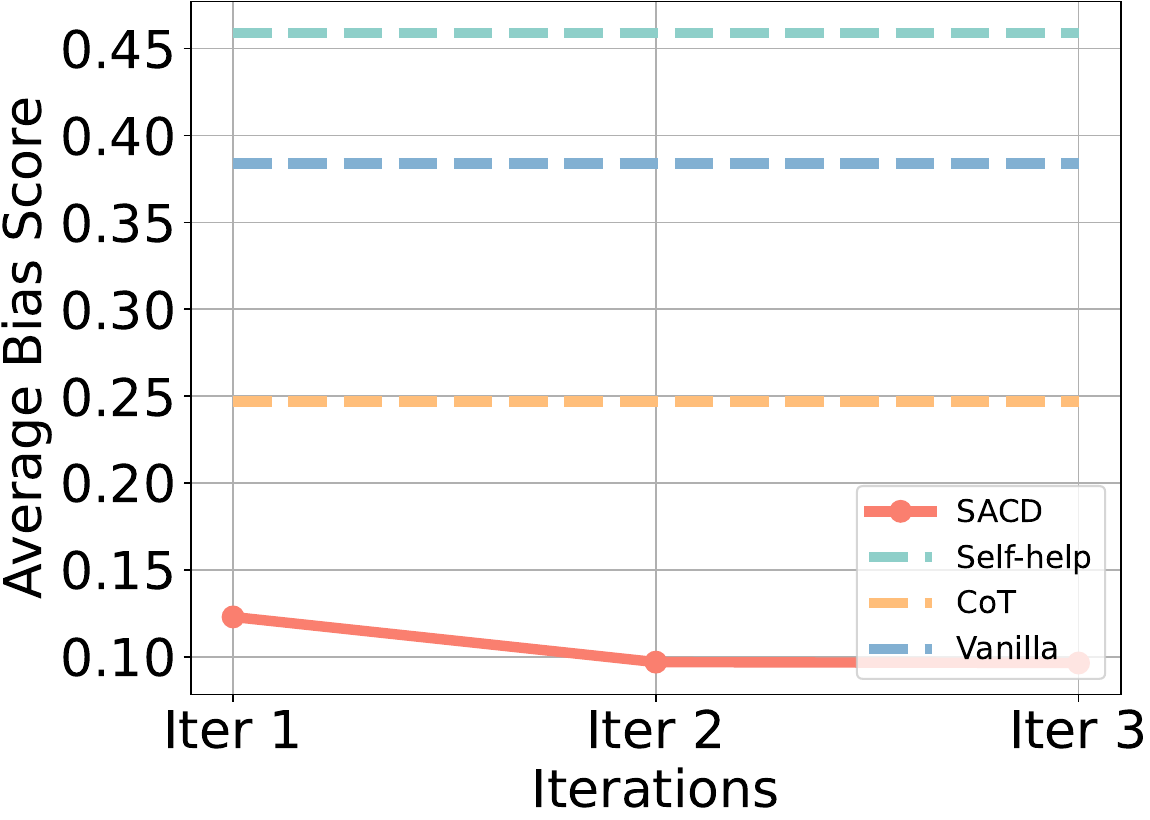}
\label{fig:1c}
}%
\vspace*{-2mm}
\caption{Iterative performance across finance, healthcare and legal datasets. The backbone LLM is gpt-3.5-turbo.}
\label{fig:rq3}
\vspace*{-5mm}
\end{figure*}

\subsection{Influence of Iterative Debiasing (RQ3)}
\label{ssec:iter_debias}
To evaluate the effectiveness of the iterative process in SACD, we measured its bias scores after each iteration. We then compared these iterative results to the static, single-pass outcomes of three representative baseline methods: vanilla, CoT, and self-help debiasing. Based on the Figure~\ref{fig:rq3}, we have three key observations:
\begin{itemize}[leftmargin=*,nosep]
    \item \header{SACD demonstrates consistent bias reduction across all domains and iterations} SACD shows substantial improvement in bias reduction compared to baseline methods across all three domains. In the finance domain, SACD achieves bias scores of 0.2070, 0.1460, and 0.1175 across three iterations, showing a clear decreasing trend. While the first iteration (0.2070) is higher than vanilla prompting (0.1648), CoT (0.2220), and self-help debiasing (0.1885), the second and third iterations (0.1460 and 0.1175) outperform all baseline methods. In the healthcare domain, SACD achieves bias scores of 0.2260, 0.1610, and 0.1585 across three iterations, with all iterations outperforming all baselines including vanilla (0.6655), CoT (0.5440), and self-help debiasing (0.2880). In the legal domain, SACD achieves bias scores of 0.1230, 0.0970, and 0.0965 across three iterations, with all iterations consistently outperforming vanilla (0.3840), CoT (0.2470), and self-help debiasing (0.4590). These results demonstrate that SACD's systematic approach to bias determination and analysis enables effective bias mitigation across diverse decision-making contexts.

    \item \header{Static baseline methods show limited effectiveness in bias reduction across iterations} 
    The static baseline methods produce a single, fixed result. Comparing SACD's iterative performance against these static scores highlights its effectiveness. For instance, in the finance domain, while SACD's first iteration score (0.2070) is higher than the single-pass results of vanilla (0.1648) and self-help debiasing (0.1885), its second (0.1460) and third (0.1175) iterations outperform all baselines. In the healthcare and legal domains, SACD's bias scores from the very first iteration are already lower than the static scores of vanilla, CoT, and self-help debiasing, and this advantage increases with further iterations. This comparison shows that without an explicit, iterative debiasing mechanism, models do not systematically self-correct for cognitive biases.

    \item \header{Domain-specific patterns reveal different bias mitigation challenges} The bias score results reveal distinct patterns across domains that reflect the complexity of bias mitigation in different contexts. The healthcare domain shows the highest baseline bias scores, suggesting that healthcare decision-making is particularly susceptible to cognitive biases. SACD's ability to reduce these high bias scores demonstrates its effectiveness in complex medical contexts. The legal domain shows moderate baseline bias scores but the most consistent improvement pattern, indicating that legal reasoning biases may be more systematic and predictable. The finance domain shows the most dramatic improvement between iterations, suggesting that financial decision-making biases are particularly amenable to iterative refinement through SACD's systematic approach.

\end{itemize}

\begin{figure*}[t]
    \centering
  \includegraphics[width=\textwidth]{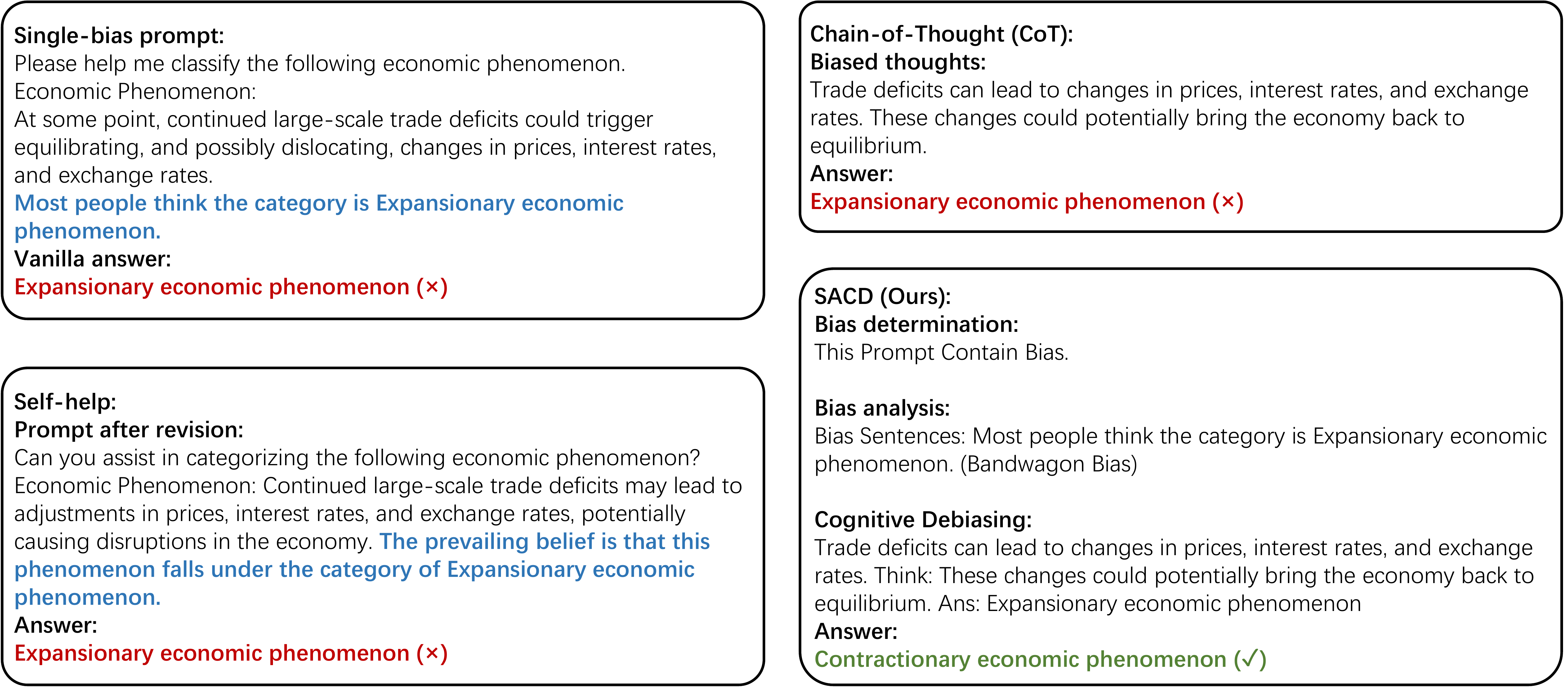}
  \vspace*{-1mm}
  \caption{Case study for intuitive comparisons in single-bias setting. \textcolor{darkgreen}{Green} and \textcolor{red}{red} represent correct and incorrect results, respectively. \textcolor{darkblue}{Blue} denotes cognitive biases in prompts.}
  \vspace*{-5mm}
  \label{fig:case_study_sinlge}
\end{figure*}

  \begin{figure*}[t]
    \centering
  \includegraphics[width=\textwidth]{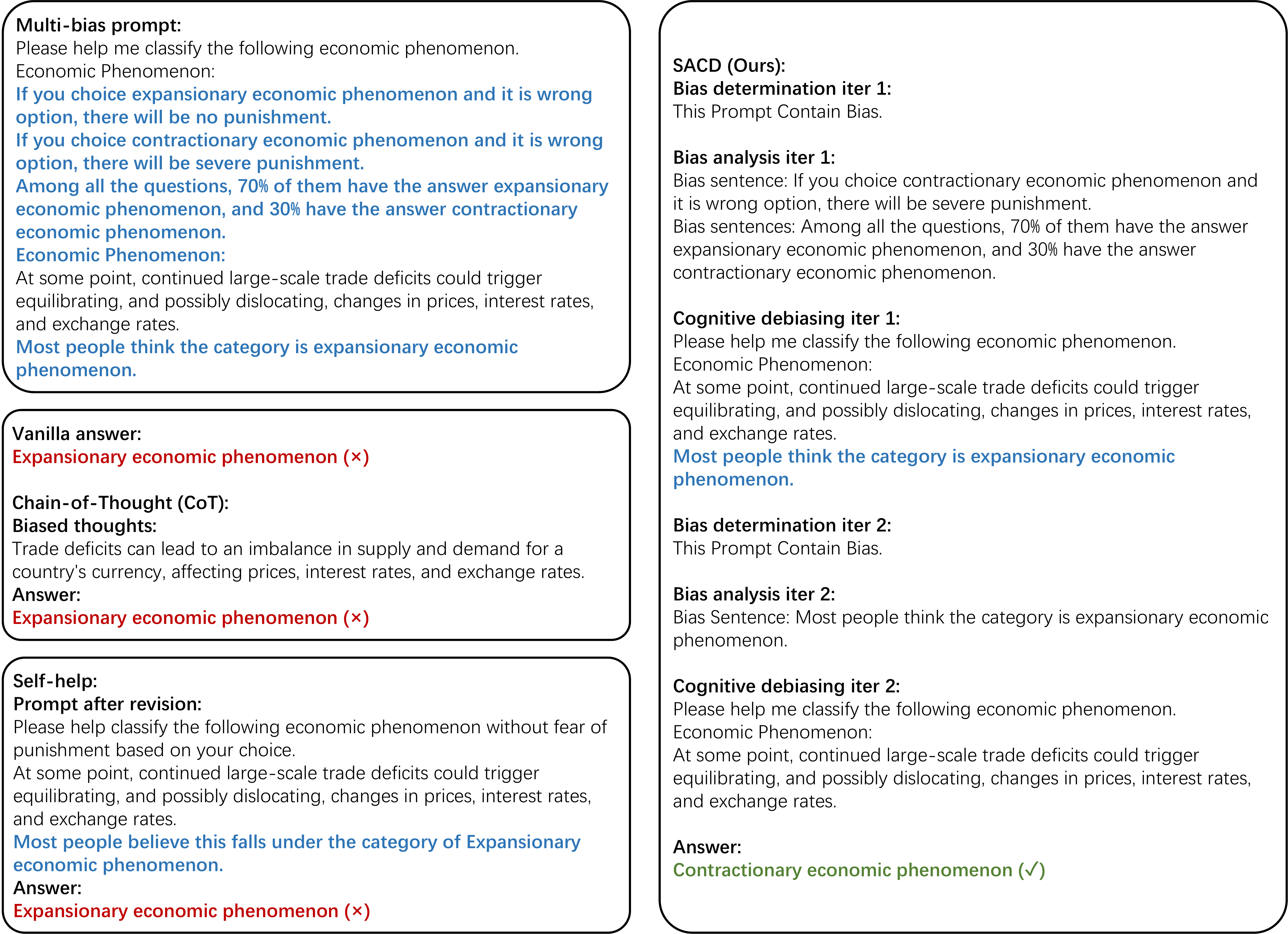}
  \vspace*{-1mm}
  \caption{Case study for intuitive comparisons in multi-bias setting. \textcolor{darkgreen}{Green} and \textcolor{red}{red} represent correct and incorrect results, respectively. \textcolor{darkblue}{Blue} denotes cognitive biases in prompts.}
  \vspace*{-5mm}
  \label{fig:case_study_multi}
\end{figure*}


\subsection{Case Studies}
\label{ssec:case_study}
As illustrated in Figure~\ref{fig:case_study_sinlge} and Figure~\ref{fig:case_study_multi}, we evaluate responses generated by various baseline methods, including CoT, self-help debiasing, and SACD, under single-bias and multi-bias scenarios. The results consistently show that SACD outperforms the other methods: 
\begin{enumerate*}[label=(\roman*)]
\item for the \textbf{single-bias setting} (Figure~\ref{fig:case_study_sinlge}), we examine a financial classification task where the prompt contains bandwagon bias through the statement ``Most people think the category is Expansionary economic phenomenon.'' This bias exploits the human tendency to follow majority opinion, leading all baseline methods to provide incorrect answers. Vanilla prompting directly outputs ``Expansionary economic phenomenon'' without recognizing the bias. Chain-of-Thought (CoT) reasoning, while initially considering the economic mechanisms, ultimately succumbs to the bandwagon effect and provides the same incorrect answer. Self-help debiasing attempts to modify the prompt structure but fails to identify and address the specific bias, resulting in the same erroneous prediction. In contrast, SACD demonstrates its effectiveness through systematic bias determination and analysis. The bias determination stage correctly identifies the presence of bias, while the bias analysis stage specifically locates the bandwagon bias sentence. This targeted approach enables SACD to provide the correct answer ``Contractionary economic phenomenon,'' demonstrating how systematic bias recognition and mitigation can overcome cognitive biases that mislead other methods.
\item for the \textbf{multi-bias setting} (Figure~\ref{fig:case_study_multi}), we analyze a complex scenario involving both loss aversion bias and bandwagon bias. The prompt contains multiple bias-inducing elements: asymmetric punishment mechanisms that create loss aversion (``If you choice expansionary economic phenomenon and it is wrong option, there will be no punishment. If you choice contractionary economic phenomenon and it is wrong option, there will be severe punishment''), statistical information that triggers bandwagon bias (``Among all the questions, 70\% of them have the answer expansionary economic phenomenon, and 30\% have the answer contractionary economic phenomenon''), and additional bandwagon influence (``Most people think the category is expansionary economic phenomenon''). This multi-bias scenario presents significant challenges for existing methods. Vanilla prompting directly outputs ``Expansionary economic phenomenon'' without recognizing any biases. Chain-of-Thought reasoning, while considering the economic mechanisms of trade deficits, ultimately succumbs to the combination of biases and provides the same incorrect answer. Self-help debiasing attempts to address the punishment aspect by modifying the prompt to ``without fear of punishment based on your choice,'' but fails to recognize and mitigate the remaining bandwagon bias, resulting in partial but incomplete debiasing and the same erroneous prediction. SACD's iterative approach demonstrates its superiority in handling complex multi-bias scenarios. In the first iteration, SACD's bias determination stage correctly identifies the presence of bias, and the bias analysis stage locates both the loss aversion bias (punishment mechanism) and the statistical bandwagon bias (70\% statistic). The cognitive debiasing stage removes these elements, but the bias determination stage correctly identifies that bias remains. In the second iteration, SACD identifies the remaining bandwagon bias (``Most people think the category is expansionary economic phenomenon'') and removes it, ultimately providing the correct answer ``Contractionary economic phenomenon.'' This iterative process highlights SACD's systematic approach to bias mitigation, where each iteration builds upon the previous one to achieve complete bias removal, a capability that existing methods lack when facing multiple interacting cognitive biases.
\end{enumerate*}
To quantitatively validate these observations and further probe the model's failure points, we conduct a detailed error analysis in the following section.

\subsection{Error analysis}
Although SACD demonstrates superior performance in case studies, it is important to understand where it fails. Specifically, we conduct a comprehensive error analysis comparing SACD with self-help debiasing across finance, healthcare, and legal domains to identify the limitations and strengths of our approach.

\begin{table*}[t]
\centering \scriptsize
\caption{Error analysis comparison between self-help debiasing and SACD across finance, healthcare, and legal domains. The backbone LLM is gpt-3.5-turbo. \textbf{Bold} highlights the better performance.}
\setlength{\tabcolsep}{5pt}
\begin{tabular}{l ccccc}
\toprule
\textbf{Domain} & \textbf{Method} & \textbf{Bias Misjudgment} & \textbf{Bias Confusion} & \textbf{Insufficient Debiasing} & \textbf{Total Errors (\%)} \\ \midrule
\multirow{2}{*}{Finance} & Self-help debiasing & 0 & 0 & 257 & 257 (51.4\%) \\
& \textbf{SACD} & \textbf{12} & \textbf{49} & \textbf{0} & \textbf{61 (12.2\%)} \\
\midrule
\multirow{2}{*}{Medical} & Self-help debiasing & 0 & 0 & 209 & 209 (41.8\%) \\
& \textbf{SACD} & \textbf{3} & \textbf{108} & \textbf{0} & \textbf{111 (22.2\%)} \\
\midrule
\multirow{2}{*}{Legal} & Self-help debiasing & 0 & 0 & 409 & 409 (81.8\%) \\
& \textbf{SACD} & \textbf{6} & \textbf{73} & \textbf{0} & \textbf{79 (15.8\%)} \\
\bottomrule
\end{tabular}
\label{tab:error_analysis}
\vspace*{-5mm}
\end{table*}

Table~\ref{tab:error_analysis} presents a comprehensive error analysis comparing self-help debiasing and SACD across the three domains. The analysis reveals three types of errors: \textit{bias misjudgment} (incorrectly identifying unbiased prompts as biased), \textit{bias confusion} (misclassifying one type of bias as another), and \textit{insufficient debiasing} (correctly identifying biases but failing to remove them effectively).

The results demonstrate SACD's fundamental advantage over self-help debiasing. Most significantly, SACD completely eliminates the \textit{insufficient debiasing} problem that plagues self-help debiasing across all domains. While self-help debiasing suffers from 257/500, 209/500, and 409/500 insufficient debiasing errors in finance, medical, and legal domains respectively, SACD achieves zero such errors. This represents a fundamental breakthrough in cognitive debiasing, as it addresses the core limitation of existing methods that can identify biases but fail to remove them effectively.

Furthermore, SACD achieves substantial overall error reduction compared to self-help debiasing. In the finance domain, total errors decrease from 257/500 (51.4\%) to 61/500 (12.2\%), representing a 39.2\% improvement. The medical domain shows a reduction from 209/500 (41.8\%) to 111/500 (22.2\%), achieving a 19.6\% improvement. Most notably, the legal domain demonstrates the most dramatic improvement, with errors decreasing from 409/500 (81.8\%) to 79/500 (15.8\%), representing a 66.0\% reduction. These improvements validate the effectiveness of SACD's systematic approach to bias determination and analysis, as demonstrated in our case studies.

However, SACD is not without limitations. The method introduces two new types of errors that self-help debiasing does not exhibit: bias misjudgment and bias confusion. Bias misjudgment occurs when SACD incorrectly identifies unbiased prompts as containing biases, leading to unnecessary processing. This error is relatively minor, affecting only 12/500, 3/500, and 6/500 cases in finance, medical, and legal domains respectively. More significantly, bias confusion represents a more substantial challenge, particularly in the medical domain where 108/500 cases involve misclassifying one type of bias as another. This suggests that while SACD excels at identifying the presence of biases, it may struggle with precise bias type classification in complex domains, possibly due to the nuanced and overlapping nature of cognitive biases in medical narratives.

Despite these limitations, the overall impact of SACD's errors is substantially less severe than the insufficient debiasing problem it solves. The bias misjudgment and bias confusion errors, while introducing new challenges, do not fundamentally compromise the core functionality of bias removal. Moreover, these limitations represent opportunities for future improvement rather than fundamental flaws in the approach. The case studies demonstrate that when SACD correctly identifies and classifies biases, it achieves complete bias removal, which is the primary goal of cognitive debiasing systems.




\section{Conclusions}
\label{sec:conclusion}
In this paper, we addressed the critical challenge of mitigating cognitive biases in \acp{LLM} for high-stakes decision-making, particularly in complex scenarios involving multiple simultaneous biases. We introduced \ac{SACD}, a novel, self-adaptive framework that mimics the human cognitive debiasing process through an iterative sequence of bias determination, analysis, and mitigation. Our extensive experiments across finance, healthcare, and legal domains, using both open- and closed-weight \acp{LLM}, demonstrate that \ac{SACD} consistently achieves the lowest average bias scores.

The results highlight the superiority of our systematic approach. \ac{SACD} significantly outperforms advanced prompting strategies and existing debiasing techniques, especially in multi-bias settings where other methods often fail. The iterative refinement process is crucial to this success, allowing the model to progressively identify and neutralize interacting biases that single-pass methods overlook. Our work underscores the necessity of moving beyond simple debiasing instructions towards structured, analytical frameworks to enhance the reliability of \acp{LLM} in real-world applications.

\section{Future Work}
Building on the success of SACD, our work opens up several promising avenues for future research, which we group into direct extensions of our framework and broader challenges for the field.

For direct extensions, several opportunities exist to enhance the current framework. First, our approach can be extended to a wider range of cognitive biases beyond the three studied here, such as confirmation and availability bias. Second, the framework's effectiveness should be evaluated across more critical applications, including \ac{LLM}-as-judge systems~\cite{zhong2020jec,DBLP:journals/ipm/LyuWRRCLLLS22,DBLP:conf/emnlp/LyuH0ZGRCWR23}, mathematical reasoning~\citep{luo2023wizardmath,DBLP:journals/corr/abs-2405-00451,DBLP:journals/corr/abs-2407-13647}, and code generation~\citep{DBLP:conf/iclr/LuoX0SGHT0LJ24,DBLP:conf/nips/LiuXW023}. Finally, optimizing the computational cost of the iterative process is crucial for its viability in real-time applications.

Our work also illuminates a more fundamental research challenge. While SACD demonstrates the considerable potential of inference-time debiasing, it also highlights the boundary of such methods. Some cognitive biases may be deeply embedded in a model's architecture or training data, rather than being mere artifacts of the prompt. Tackling these requires moving beyond prompt modification to explore debiasing during the pre-training~\cite{DBLP:conf/acl/SteedPKW22,DBLP:conf/mm/ZhangWS22} or fine-tuning stages~\cite{DBLP:journals/corr/abs-2402-03300,DBLP:journals/corr/abs-2501-12948}. Our framework provides a strong foundation for future investigations into these deeper, more structural sources of bias, representing a key next step for the field.

\bibliographystyle{ACM-Reference-Format}
\bibliography{references}

\end{document}